# Responsible and Regulatory Conform Machine Learning for Medicine: A Survey of Challenges and Solutions

**EIKE PETERSEN**[1,2], **(Member, IEEE), YANNIK POTDEVIN**[3], **ESFANDIAR MOHAMMADI**[4], **STEPHAN ZIDOWITZ**[5], **SABRINA BREYER**[2], **DIRK NOWOTKA**[3], **SANDRA HENN**[2], **LUDWIG PECHMANN**[6], **MARTIN LEUCKER**[6,7], **PHILIPP ROSTALSKI**[2,8], **(Member, IEEE), AND CHRISTIAN HERZOG**[2], **(Member, IEEE)**

[1]DTU Compute, Technical University of Denmark, 2800 Lyngby, Denmark
[2]Institute for Electrical Engineering in Medicine (IME), Universität zu Lübeck, 23562 Lübeck, Germany
[3]Department of Computer Science, Kiel University, 24143 Kiel, Germany
[4]Institute for IT Security (ITS), Universität zu Lübeck, 23562 Lübeck, Germany
[5]Fraunhofer Institute for Digital Medicine MEVIS, 28359 Bremen, Germany
[6]UniTransferKlinik Lübeck GmbH, 23562 Lübeck, Germany
[7]Institute for Software Engineering and Programming Languages (ISP), Universität zu Lübeck, 23562 Lübeck, Germany
[8]Fraunhofer Research Institution for Individualized and Cell-Based Medical Engineering (IMTE), 23562 Lübeck, Germany

Corresponding author: Eike Petersen (ewipe@dtu.dk)

This work was supported in part by the German Federal Ministry for Economic Affairs and Climate Action (BMWK) through the AI Space for Intelligent Health Systems (KI-SIGS) Project; in part by several universities, research institutions, and private companies in northern Germany; and in part by the Academy of Sciences and Humanities, Hamburg (by paying the open access fees). The work of Yannik Potdevin, Sabrina Breyer, and Ludwig Pechmann was supported by the KI-SIGS Project.

**ABSTRACT** Machine learning is expected to fuel significant improvements in medical care. To ensure that fundamental principles such as beneficence, respect for human autonomy, prevention of harm, justice, privacy, and transparency are respected, medical machine learning systems must be developed responsibly. Many high-level declarations of ethical principles have been put forth for this purpose, but there is a severe lack of technical guidelines explicating the practical consequences for medical machine learning. Similarly, there is currently considerable uncertainty regarding the exact regulatory requirements placed upon medical machine learning systems. This survey provides an overview of the technical and procedural challenges involved in creating medical machine learning systems responsibly and in conformity with existing regulations, as well as possible solutions to address these challenges. First, a brief review of existing regulations affecting medical machine learning is provided, showing that properties such as safety, robustness, reliability, privacy, security, transparency, explainability, and nondiscrimination are all demanded *already* by existing law and regulations — albeit, in many cases, to an uncertain degree. Next, the key technical obstacles to achieving these desirable properties are discussed, as well as important techniques to overcome these obstacles in the medical context. We notice that distribution shift, spurious correlations, model underspecification, uncertainty quantification, and data scarcity represent severe challenges in the medical context. Promising solution approaches include the use of large and representative datasets and federated learning as a means to that end, the careful exploitation of domain knowledge, the use of inherently transparent models, comprehensive out-of-distribution model testing and verification, as well as algorithmic impact assessments.

## I. INTRODUCTION

The potential of modern machine learning techniques to significantly improve clinical diagnosis and treatment, and

The associate editor coordinating the review of this manuscript and approving it for publication was Wenbing Zhao.

to unlock previously infeasible healthcare applications, has been thoroughly demonstrated [1]–[3]. At the same time, the risks posed by the application of medical machine learning (MML) have become apparent. Critical obstacles that have been identified include *non-robust models* that do not generalize to broader patient cohorts or different

  

settings, exaggerated performance claims based on *inadequate model evaluation*, a *lack of transparency* regarding the reasoning behind model predictions, *privacy* challenges, and unintended *discriminatory biases* against certain patient groups [4]–[8]. Currently, the Partnership on AI's "Artificial Intelligence Incident Database"[1] (which is neither limited to medical systems nor machine learning-based systems) lists more than 1200 reported "situations in which AI systems caused, or nearly caused, real-world harm," illustrating that these technical challenges have real-world consequences [9]. Notable examples of reported problems with *medical* machine learning systems include

- a system diagnosing hip fractures mainly based on patient traits and hospital process variables, almost regardless of the patient's X-ray recording [10],
- poor generalization performance of a chest radiograph-based pneumonia detection system from one recording site to others [11],
- surgical skin markings influencing the decision-making of a system designed for melanoma recognition [12],
- improper performance evaluation methodology leading to widely exaggerated performance claims for clinical prediction models [8],
- disparities in true positive rates of chest X-ray diagnosis systems between patients of different sex, age, race, or insurance types [13], and
- racial bias in an algorithm that computes a health risk score and is used to assign healthcare system resources in the US [14].

Figure 1 shows a schematic overview of some of the many places in the MML workflow in which risks arise, as further discussed in this article.

Besides these technical challenges, creating a regulatory environment that does not stifle innovation, yet addresses potential sources of harm, will be crucial to facilitate a prosperous future for MML [5], [7], [15]–[17].

### A. BACKGROUND

A great many reviews and surveys have been published about the state of machine learning in medicine. General and rather practical introductions to the topic [18], systematic literature reviews of existing work [19] and reviews of the state of regulations concerning MML [17], discussions of the role that machine learning can and should play in medical practice [3], [20], and reflections on the pertinent ethical questions [21] have been published. Qayyum *et al.* [22] provide a survey of secure and robust machine learning for healthcare, discussing security and privacy challenges as well as potential solutions. Ghassemi *et al.* [23] and Aristidou *et al.* [24] both discuss challenges to the widespread adoption of ML-based systems in clinical practice and avenues for their solution. What can we contribute to this vast body of existing work?

In the wider field of general machine learning, there is currently a significant gap between high-level ethical and regulatory requirements on the one hand and their practical implementation on the other hand [25]–[27]; this has been called the "principles–to–practices gap" [28], which may also be partially responsible for the slow adoption of ML-based devices in clinical practice [3], [24]. To date, only few documents are available that provide practical technical guidelines aiding engineers in responsibly implementing ML applications [29], [30], and none of them discuss the specific challenges posed by medical applications. The FDA's proposed regulatory framework for modifications to AI/ML-based software as a medical device [31] mentions the requirement to follow "good machine learning practices (GMLP)" but does not provide further information about such best practices. While, in a follow-up publication, the FDA, Health Canada, and the United Kingdom's Medicines and Healthcare products Regulatory Agency have now proposed "guiding principles" for the *development* of such best practices [32], there is still no detailed practical guideline available. Part of the reason for the relative lack of practical guidance is the fact that — despite a flurry of research activity in these areas in recent years — the research community is only beginning to propose solutions to many of the relevant technical challenges, with new technical problems surfacing frequently. Moreover, it appears unlikely that there will ever be a universal procedure to successfully implement a safe, robust, reliable, privacy-preserving, transparent, and fair machine learning system for a given application. This is not to diminish the immense progress made in recent years in areas such as adversarial robustness, explainability, algorithmic fairness, differential privacy, and uncertainty quantification, among many other relevant fields. However, none of these challenges should be considered solved. Moreover, a synthesis of many of these diverse technical advances into practical technical guidance is lacking, particularly concerning the unique requirements of medical ML.

Finally, it has also been noted [26] that most AI white papers are "top-down", i.e., written by governmental institutions and policymakers who have little contact with the actual development and use of AI systems. There is a perceived lack of "bottom-up" initiatives on practical AI ethics, driven by, e.g., hands-on AI developers and the people affected by the AI system [26], although there are notable counter-examples [33]. Here, we aim to provide such a bottom-up perspective on the technical aspects of responsible and regulatory conform medical machine learning.

### B. SCOPE AND PURPOSE OF THIS DOCUMENT

The purpose of this document is to provide an overview of and practical guidance concerning the challenges and potential solutions involved in creating medical ML systems responsibly, taking into account the current and likely future regulatory landscape on the one hand and current technical best practices on the other hand. Our main contributions are as follows:

---

[1]See https://incidentdatabase.ai/

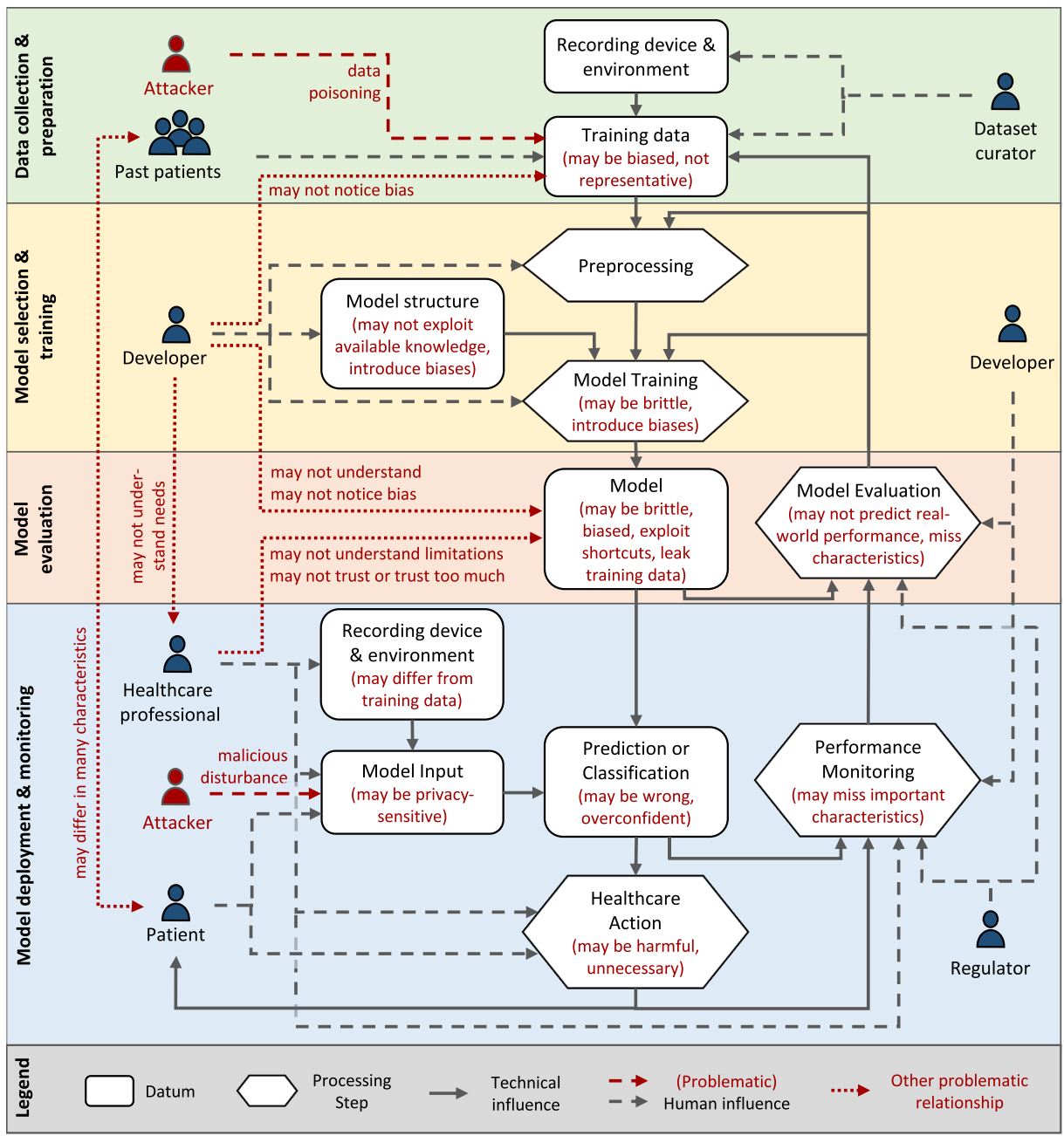

**FIGURE 1.** A schematic overview of the medical machine learning workflow throughout the model's life cycle, some of its stakeholders, and some of its risks. The risks and possible solutions to mitigate them are discussed throughout this document.

1) We show that far-reaching requirements are already placed upon medical ML systems by today's existing regulations, and that these requirements appear very likely to be further substantiated by future regulatory developments.

2) We provide a thorough discussion of the key obstacles that need to be overcome in order to design systems that fulfill the requirements posed by regulations and ethical considerations, as well as potential technical and non-technical approaches to their solution and their respective merits and drawbacks. In essence, we attempt to provide practical guidance on how to implement responsible machine learning for medical applications.

The different sections of this document are mostly self-contained, hopefully making them accessible to readers who are only interested in the regulatory discussion or specific technical challenges.

Lastly, several notes on the scope of this document. First, we do *not* aim to provide precise technical best-practice recommendations, simply because research on these challenges is still young and evolving quickly. Instead, we provide an expository overview of each technical challenge, why it arises, how it may instantiate in a medical context, and promising approaches to its solution. Second, we will not address *dynamical* MML systems here. By dynamical MML systems, we refer to both continuously learning MML systems as well as ML-based closed-loop control of physiological systems (without any human decision-making in the loop), such as automatic insulin delivery systems [34], [35] and closed-loop neural interfaces [36]. Third, we will *not* address the social implications of developing and deploying an MML system, the question of whether such a system should be built at all, questions of data governance, nor the implementation of inclusive and participatory development processes. A broad overview of these important ethical, societal and governmental challenges, as well as proposed solutions, can be found in recent guidance by the WHO [7]. Here, we will focus on the remaining *technical and procedural challenges* and assume that a decision has already been made to realize a particular healthcare functionality using a machine learning system. Thus, the question that remains is this: how to develop such a system responsibly and in a way that ensures regulatory approval?

### C. METHODOLOGY

We conduct a narrative review of the challenges and potential solutions towards achieving responsible and regulatory conform machine learning for medicine. To this end, in an initial literature search, the five key themes that make up the remainder of this article were identified: regulations and responsibility firstly, and then the four technical challenges outlined below. In the next step, expert researchers in those five topics were identified and invited to join the project as collaborators. Drafts of the individual sections were then developed by the chosen experts in close collaboration with the first author. In a final integration step, the different parts were unified and harmonized, and the overall findings synthesized. Due to the qualitative nature of our study, formal criteria for literature inclusion or exclusion were not employed.

### D. OUTLINE

The following section II-A will provide a brief overview of regulations concerning medical ML, as well as the main challenges to overcome in order to obtain regulatory approval for a medical ML system. Readers who are more interested in the technical discussion of obstacles and solutions than in the regulatory landscape may simply skip this part. From this overview, four key requirements for the realization of responsible and regulatory conform medical ML emerge:

1) safety, robustness & reliability,
2) privacy & security,
3) transparency & explainability, and
4) algorithmic fairness & nondiscrimination.

Figure 2 illustrates the wealth of regulations and ethical fields affecting MML, as well as the aforementioned key requirements for responsible and regulatory conform MML.

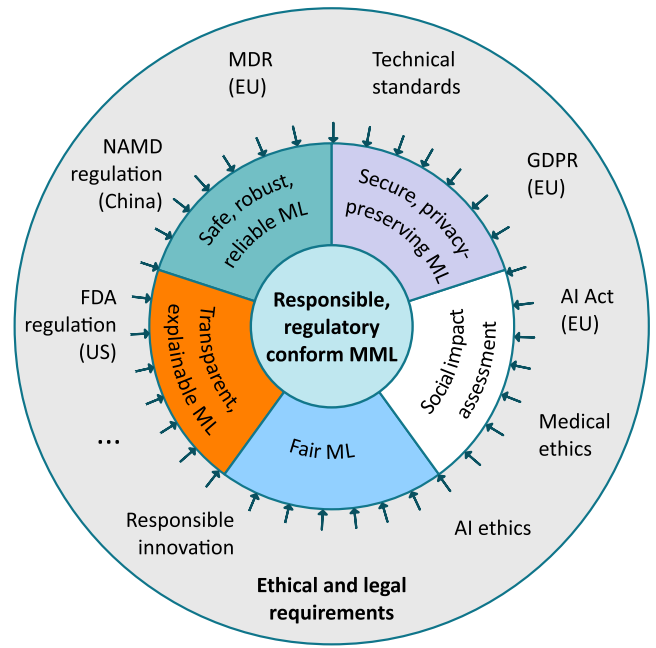

**FIGURE 2.** Medical machine learning lies at the intersection of medicine and machine learning and is subject to regulations and ethical considerations regarding both fields. The key requirements that emerge from all of these are safety, robustness, reliability, privacy, security, transparency, explainability, and fairness. In order to perform *responsible* MML, social impact assessment is another key requirement, which is not addressed in this document.

In section III to section VI, we then proceed to discuss the current state of the art regarding solutions to these four challenges, always taking particular account of the peculiarities of medical ML applications. In section VII, we return from the technical details to discuss the high-level themes that emerge, as well as important open problems and future challenges (both technical and regulatory).

## II. REGULATIONS AND RESPONSIBILITY

Medical products based on machine learning algorithms are already receiving regulatory approval based on current medical device regulations [37], [38] which impose substantial requirements on the development process, the final MML product, and post-market surveillance. Furthermore, broad regulations such as the EU's general data protection regulation (GDPR) or nondiscrimination law equally apply to MML, of course [39]–[42]. The exact consequences of these existing regulations for MML applications are far from fully developed. At the same time, the regulatory landscape concerning general AI/ML systems (thus including MML systems) is evolving rapidly, further exacerbating regulatory uncertainty. Dozens of countries and organizations have published general principles for ethical AI [27], and regulatory approaches, as well as technical standards, are currently under heavy development [7], [43]–[49]. Due to the preliminary, fast-changing nature of these efforts, it is currently uncertain which standards and regulations companies will have to comply with in order to obtain regulatory approval for MML systems in years to come. It appears to be clear, however, that besides the already existing medical device regulations (MDR, FDA 21 CFR and IEC 62304, to name just a few examples), MML will need to follow general ethical principles for the use of AI such as those put forth by the EU expert commission on AI [50], [51], the WHO [7], the OECD [52], or the Chinese ministry of science and technology's national new generation AI governance expert committee. Although there is currently an abundance of proposed guidelines for ethical AI, most of these proposals converge on a small set of core values: beneficence, respect for human autonomy and dignity, prevention of harm, justice and fairness, privacy, transparency and explicability, and responsibility and accountability [25], [53]–[55]. Recently, the WHO has released a comprehensive guidance document concerning the governance and ethics of AI for health [7], which also agrees with the above key ethical principles. An MML application that implements these core values to a satisfactory degree can thus be considered subject to low regulatory risks with regard to future regulatory changes.

*Responsibility*, of course, demands more than just compliance with existing law and regulations. Firstly, regarding those properties that *are* subject to existing regulations, such as, e.g., patient safety and nondiscrimination, it may be perfectly feasible to design a system that successfully receives regulatory approval yet is not as safe or non-discriminatory as it could (and maybe should) be. This discrepancy may be due to the rapidly evolving body of knowledge about the underlying technology and its associated risks (leaving regulatory processes trailing behind), a lack of technical precision in regulations, less than rigorous compliance evaluations, or the influence of lobbying on regulations and standards [56]. Thus, from the perspective of a responsible developer or manufacturer, it may not be morally sufficient to aim for the minimum level of technical and procedural safeguards that enables receiving regulatory approval.

Secondly, responsibility encompasses more dimensions than are covered by current regulations, most importantly *social* and *moral* dimensions [7], [57]. Quoting Dignum [58], responsibility is "not just about making rules to govern intelligent machines; it is about the whole sociotechnical system in which the system operates, and which encompasses people, machines and institutions." In the context of machine learning for medicine, this includes the wider implications of a new ML-based product for the patient and his or her relatives, the hospital and its stakeholders, the healthcare system, and the society at large. As an example, Elish and Watkins [59] study the impact of deploying "Sepsis Watch", a deep learning-based system for predicting a patient's risk of developing sepsis, on the daily work of hospital nurses. They find that the introduction of the MML system presents new challenges for these nurses and significantly changes their responsibilities and daily work. As a *disruptive* innovation, the new system does not fit into established hospital procedures, hierarchies, and communication rituals, thus requiring a significant amount of "repair work" (mainly by the nurses tasked with implementing the system) to be integrated into the sociotechnical hospital system successfully [59]. Moreover, it challenges physician autonomy (thereby complicating successful communication of the computed risk scores) and demands the integration of a new type of information (an inscrutable risk score computed by the system) into established clinical decision-making processes [59]. Stepping back from this particular example, there are many ethical discussions to be led about whether some particular functionality should be realized using an artificially intelligent system *at all*, or under which circumstances [7], [26], [60]–[63]. To address these and similar challenges, various frameworks for algorithmic impact assessments have been proposed [7], [33], [64]–[67]. On the other hand, one may also argue that there is a moral responsibility to make improved treatments available, if possible and if there are no opposing moral reasons not to do so. We will *not* address these questions in our survey. Instead, we will assume that a decision has already been made (responsibly) to realize a particular healthcare functionality using a machine learning system.

## A. EXISTING REGULATIONS

A brief review of laws, standards, and guidelines influencing MML development as of today (cf. fig. 2) shows that there is very little ML-specific regulation as of today. Several jurisdictions have issued regulation of specific ML-enabled technologies such as facial recognition and autonomous driving, but neither general regulation of ML systems nor of medical ML systems [68]. Thus, currently, ML-based medical devices are regulated based on classical, non-ML-specific medical device regulation such as the MDR in the EU and the Code of Federal Regulation Title 21 in the US [17], [37], [69]–[71]. Additionally, broad, non-medicine-specific regulations such as the EU's general data protection regulation (GDPR) [42], and anti-discrimination law, of course, apply. In the context of medical device regulations, MML systems are generally interpreted as software (the model and its implementation), of which large parts are automatically generated by other software (the training procedure) in a data-dependent way [31], [72]. As all three components (training data, training procedure, final model implementation) affect the resulting MML system, all of them are subject to regulatory scrutiny. Currently, the path to regulatory approval for medical ML systems strongly depends on the particular interpretations of existing law pursued by the auditing institution, and, as a consequence, evaluations are performed with varying methodologies and rigor [42], [69], [73]. Nevertheless, a considerable number of devices have already received regulatory approval based on these regulations [37], [69]. Thus, the question arises: which requirements on MML systems can be derived from these classical medical device regulations?

Medical device regulations require manufacturers to demonstrate the beneficence, functionality and safety of their products, demanding, among many other requirements, precise specification of an *intended use*, *lifecycle* *management* including comprehensive *post-market surveillance*, *risk management*, and *verification and validation* [17], [71], [73]. For medical device software, software as a medical device (SaMD) or health software, IEC 62304 (*Medical device software – Software life cycle processes*) and IEC 82304 (*Health software*) specify the most critical requirements and best practices to follow, in addition to the rather general IEC 60601-1 (*Medical electrical equipment - Part 1: General requirements for basic safety and essential performance*) [74]–[77]. One essential requirement concerns the deployment of a *quality management system (QMS)* (following, e.g., the ISO 13485 [78] in Europe or FDA 21 CFR Part 820 in the U.S. [79]) as an essential step to ensure that the final product is safe, effective, and efficient [80]. Such a QMS requires the documentation of the whole product lifecycle, including the product planning stage, design and development, verification and validation, as well as deployment, maintenance, and disposal [80]. In addition to the use of a QMS, proper *risk management* (typically following ISO 14971 [81]) is another core requirement that must be pursued throughout the whole lifecycle, and that affects every stage of the development process [71], [73], [74], [81], [82]. A system's intended use must be specified precisely, including, e.g., the medical indication, the treated body part, the target patient population, the intended user (clinical doctor, nurse, patient, patient's family?), and the intended usage environment [81], [83], [84]. Based on the intended use, the product is assigned a *safety class* that determines the level of further safety requirements [17], [79], [81], [84], [85]. (Notably, any system that is intended to monitor physiological processes is assigned risk class IIa or higher according to the MDR, indicating that most MML systems will likely fall into this or a higher risk class [17].) Potential risks — arising during the intended use as well as during "reasonably foreseeable misuse" [84] — must be identified and mitigated, either by reducing their likelihood of occurrence or their severity [81]–[84]. Risk mitigation measures may include technical improvements to the system itself, usability engineering measures (such as described by IEC 62366-1 [83]) like mandatory user training and careful UI design, as well as modifying the system's intended use [77], [81], [83], [84]. Finally, extensive *post-market surveillance* activities are also required by the EU's MDR [84, chapter VII], the FDA [86], and various risk management standards [78], [81], [87]. As an example, the EU's MDR demands post-market surveillance by the manufacturer to continuously monitor and report on the safety, usability, and beneficence of the product, identify and report on an unexpectedly increased incidence or severity of faults, analyze any severe incidents, and perform appropriate field safety corrective actions.

Risk management is, thus, a central paradigm of medical device regulation. Interestingly, many of the well-known technical challenges with machine learning systems can be understood as risks within a risk minimization framework: non-robust models that react to spurious input changes, susceptibility to adversarial attacks, a lack of transparency to the users, and biased decision-making all pose risks to the patient, the hospital, and, potentially, the healthcare system as a whole. Accordingly, technical measures to ensure the safety, robustness, reliability, privacy, security, transparency, and fairness of the MML system all represent risk mitigation strategies that are (to some degree, at least) required by already existing regulations [88]. To further substantiate this claim, in the proposed regulatory framework on AI recently put forth by the EU commission [49], two of the four stated objectives of the policy proposal relate to the effective governance and enforcement of "existing law on fundamental rights and safety requirements applicable to AI systems". (The other two concern legal certainty for manufacturers and the development of a single, harmonized market.) In this context, it is also relevant to note that the MDR broadly requires "taking account of the generally acknowledged state of the art" [84] regarding risk control measures and safety principles. In current practice, "state of the art" is interpreted conservatively as "applicable standards and guidance documents, information relating to the medical condition managed with the device and its natural course, benchmark devices, other devices and medical alternatives available to the target population" [89].

Regarding the *verification and validation* of MML systems, the regulations distinguish between the clinical applicability of the product, the validation of the software used in the medical product itself, and the validation of the *tools* used to develop that software. Based on the safety class of the designated product, the final product (including any machine learning model) must pass various stages of clinical validation [85]. Most of the classical machine learning validation activities fall under this category. In addition, the software used to implement the trained model (often a subset of a public ML library) must be tested for its correctness as it is considered a *software of unknown provenance (SOUP)* according to IEC 62304 [72]. What many ML practitioners may be unaware of, however, is that the tools used to develop that software must *also* be validated to obtain regulatory approval — this includes, in particular, any machine learning libraries or frameworks that have been used to label or preprocess training data or to perform the actual model training [72]. This *tool validation* follows a classical software testing scheme: the desired software functionality is specified, and test cases are defined or automatically generated that validate the correctness of the software.[2] In particular, it must be demonstrated that, given the training dataset and the selected model structure and hyperparameters, the training process has correctly identified an "optimal" set of model parameters [72]. This demonstration may involve classical estimation error metrics. Again, notice that this tool validation, and the requirements placed thereupon, differs from a validation of the performance of the trained model: these two elements of the overall validation efforts serve complementary purposes. Finally, note that *transfer learning*, i.e., the use of pre-trained models (often pursued to reduce the necessary amount of domain-specific training data), entails complex requirements on the validation procedure [90].

---

[2]Importantly, only those functions of a tool or library that are used during the development process must be validated.

Since 2018, the EU's general data protection regulation (GDPR) imposes strong additional constraints on data processing systems, including (medical) machine learning applications [42], [88], [91], [92]. (While the GDPR is EU law, it famously has affected companies worldwide [93].) Firstly, the "right to an explanation" has been the subject of intense academic debate [91], [94], [95]. It has spurred a sprint of research in explainability methodology [96] and has been translated into national legislation in several EU countries [97]. GDPR article 13, among other relevant paragraphs, requires a user to be provided with "meaningful information about the logic involved" in automated decision-making. Based on the GDPR, the UK Information Commissioner's Office (ICO) has recently issued comprehensive guidance on explaining decisions made by AI systems [98]. While the exact extent to which the GDPR mandates a "right to an explanation" is currently unclear, it appears evident that black-box AI without any further explanation is *not* GDPR-compliant. Secondly, GDPR article 22(3) grants a (closely related) "right to contest the decision", which opens up questions regarding the interplay of contestability, model explainability, and transparency [92]. Thirdly, GDPR article 22(4) grants a "right to nondiscrimination" with regards to automated data processing and profiling, explicitly forbidding discriminatory use of sensitive attributes such as, among others, racial or ethnic origin, religion, and political opinion [41], [88], [91], [99]. Again—the exact requirements this places on machine learning applications are currently unclear [91], but completely ignoring bias and discrimination issues is undoubtedly not GDPR-compliant [88]. Fourthly, GDPR article 35(7) requires a Data Protection Impact Assessment (DPIA), which should contain "1. a systematic description of the envisaged processing operations and the purposes of the processing, [. . .]; 2. an assessment of the necessity and proportionality of the processing operations in relation to the purposes; 3. an assessment of the risks to the rights and freedoms of data subjects [. . .]; and 4. the measures envisaged to address the risks." It has been argued that such a DPIA in the sense of the GDPR can serve as a form of algorithmic impact assessment, although it lacks essential elements such as mandatory public disclosure of the DPIA and explicit mandatory stakeholder involvement [67]. Finally, another rather obvious consequence of the GDPR concerns the maintained privacy of the training data, which may necessitate the use of privacy-preserving ML techniques as more and more information extraction attacks become known in the literature [88], [100]–[103].

### B. FUTURE REGULATORY CHANGES
Based on their previous white paper [104], [105] and the report of the high-level expert group on AI [50], in April 2021

the EU commission has published the first concrete proposal on regulating AI-based applications [49]. The proposed regulation is derived from four objectives: to ensure that AI systems are safe and respect existing law on fundamental rights and values (including, e.g., nondiscrimination law), to ensure legal certainty, to enhance the effective enforcement of already-existing law on fundamental rights and safety requirements, and to prevent market fragmentation due to opposing regulations. Following the proposal, applications would be classified as falling into different risk categories, with most medical applications being classified as "high-risk AI systems". A high-risk classification would entail a list of legal requirements "in relation to data and data governance, documentation and recording keeping [sic], transparency and provision of information to users, human oversight, robustness, accuracy and security" [49]. Notable requirements placed on high-risk applications include

- the use of a risk management system that "shall consist of a continuous iterative process run throughout the entire lifecycle" (article 9) as well as an appropriate quality management system (article 17),
- quality criteria regarding the training data as well as their governance and management processes (article 10),
- that "their operation is sufficiently transparent to enable users to interpret the system's output and use it appropriately" (article 13), and
- that they achieve "an appropriate level of accuracy, robustness and cybersecurity, and perform consistently in those respects throughout their lifecycle" (article 15).

To retain flexibility, the precise technical solutions to achieve compliance with these requirements are left to technical standards (to be developed) and other guidelines. Outside the EU, essentially all larger regulatory bodies have issued white papers indicating their inclination to set up AI/ML-specific regulation in the near future [68]. These include the FDA's "Artificial Intelligence/Machine Learning (AI/ML)-Based Software as a Medical Device (SaMD) Action Plan" [43], emphasizing a need for improved methods to evaluate and address algorithmic bias and a need for transparency regarding the training data, the model structure, and evidence of the model's performance. Similarly, the Chinese State Council has proposed a "New Generation Artificial Intelligence Development Plan" (AIDP) [106]. Aside from these coming hard law changes, many technical standards and guidelines on subjects such as ML robustness, transparency, and bias mitigation are currently under development and will be available soon. Notable examples include the work of the ISO/IEC JTC 1/SC 42 [46], [47], the IEEE P7000 family of standards [45] and the P2801 and P2802 standards [107], [108], and the UK information commissioner's office (ICO) guidance on auditing AI algorithms [88] and explaining decisions made by AI systems [98].

### C. SUMMARY: REGULATIONS AND RESPONSIBILITY
To summarize, current regulation requires comprehensive risk assessment and mitigation, quality management, data privacy, nondiscrimination, and some degree of transparency about the reasoning behind an automated decision, for any medical product. Within the context of medical ML, the pursuit of properties such as safety, robustness, privacy, security, transparency, and freedom from discrimination against patient groups can thus be framed as *risk mitigation strategies* [88], [104], [109]. These requirements are legally binding already *today*, with future standards and regulatory changes only aiming to further clarify and substantiate them [49], [73], [88]. Considering these requirements during development today, although their current evaluation by regulatory bodies appears non-comprehensive [38], thus seems advisable not only from an ethical responsibility perspective but also from a legal point of view. Moreover, it aids in building expertise in these crucial topics that will continue to rise in importance swiftly. To this end, the following sections of this document will provide brief overviews of the technical challenges and possible solutions to achieve safe, robust, reliable, secure, privacy-preserving, transparent, and fair medical ML. A summary of these obstacles and approaches can be found in table 1.

## III. SAFETY, ROBUSTNESS AND RELIABILITY

Medical technology directly or indirectly affects patient health and therefore must adhere to rigorous safety requirements. The classical approach to developing a safe medical device is to follow a *risk-based approach*, cf. section II-A — this requires the identification of possible faults and the quantification of their likelihood of occurrence, as well as the risks that may arise from them [82]. If unacceptable risks are identified, actions must be taken to prevent and control (mitigate) the faults causing these risks [82]. For classical medical hardware devices, developers must consider safety measures such as radiation protection, fire protection, or protection against mechanical hazards. The safety aspect that is most significantly affected by the use of machine learning in a medical product, however, is *functional safety*.

A functionally safe system reliably performs its intended function: if faults with a hazardous effect occur, a functionally safe system is able to recover a safe state nevertheless [263].[3] To achieve functional safety, a system must thus i) prevent any faults in advance or, where this is infeasible, ii) reliably detect, and iii) control them. Two types of faults are usually differentiated [263]: systematic faults and random faults. A *systematic fault* is defined as a "failure, related in a deterministic way to a certain cause, which can only be eliminated by a modification of the design or of the manufacturing process, operational procedures, documentation or other relevant factors" [263]. Systematic faults may arise from construction errors, software bugs, or human operator mistakes. A *random fault*, on the other hand, is defined as "a failure occurring at a random time, which results from one or more degradation mechanisms" [263]. In classical hardware components, these are often a consequence of aging processes that unavoidably lead to hardware failures. Their rate of occurrence can be quantified and judged regarding its acceptability. The IEC 61508 [263] generally assumes that software faults are systematic and not random, which would signify that faults of an MML system are all to be considered systematic. However, the standard is currently not explicit in its characterization of machine learning systems and their failures, and neither are other standards concerning functional safety [264].[4] Concerning medical ML systems, problems such as biased datasets and fragile models represent examples of systematic errors: they result from an imperfect model design process, akin to classical software bugs. Transparency and explainability, discussed in depth in section V, serve to reduce the likelihood of systematic errors occurring by enabling the developer to inspect and analyze the system in detail. Once the system is deployed, transparency and explainability *to the user* facilitate the early detection of faults of an MML system by a human operator. section VI is concerned exclusively with methods to prevent systematic biases occurring in the learned model.

In the following discussion of barriers and measures to achieve functional safety of MML systems, we distinguish between the three closely related properties of *robustness*, *reliability* and *safety*, for which we employ the definitions also used by Borg *et al.* [265]:

**Robustness** denotes "the degree to which a component can function correctly in the presence of invalid inputs or stressful environmental conditions" [266].

**Reliability** denotes "the probability that a component performs its required functions for a desired period of time without failure in specified environments with a desired confidence" [267].

**Safety** denotes "freedom from unacceptable risk of physical injury or of damage to the health of people, either directly or indirectly, as a result of damage to property or to the environment" [263].

We (informally) call a machine learning method *robust* if it yields similar models when applied to training data drawn from similar distributions.[5] Consider, for example, an ML method producing an artificial neural network (ANN) to detect lung cancer based on X-ray images. Now suppose that two training sets of X-ray images of human lungs have been recorded from the same subjects with X-ray scanners from two different manufacturers *A* and *B*, with the scanner of manufacturer *A* yielding noisier images than the scanner of manufacturer *B*. We call the ML method *robust* if the two classifiers resulting from training on the two datasets are similar. A closely related yet subtly different notion is the one of robust ML *models* (as opposed to robust ML *methods*, comprising the whole ML tool-chain which yields an ML model). In accordance with various formal definitions of (local) robustness proposed in the literature [269], we informally denote a machine learning model to be robust if it provides similar predictions for similar inputs. Returning to the above example, we would call a classifier robust if it classified images taken from the same patient similarly, regardless of the scanner used for the recording. It is well known (and intuitively makes sense) that the robustness of a model against variations along a certain dimension is closely linked with the model's response surface being *smooth* along that dimension [269], [270]. Thus, robustness-enforcing methods can be understood as biasing the model towards smoothness along certain dimensions. Notice, however, that the maximally robust model is the one that returns a constant output regardless of the input — thus, it immediately becomes apparent that robustness on its own cannot serve as an optimization target. Robustness can therefore only be considered beneficial insofar as it serves to increase the *reliability* of the model, i.e., its tendency to perform well for most of the target data [271].

---

[3]The following discussion of functional safety is based on IEC 61508:2010 [263].

[4]Standardization efforts are currently in progress, e.g., at the ISO/IEC JTC 1/SC 42: https://www.iso.org/standard/81283.html

[5]This informal definition is in the spirit of classical robust statistical estimation theory, see, e.g., Daszykowski *et al.* [268].

**TABLE 1.** Building blocks for responsible machine learning in medicine. A check mark (✓) indicates that a solution approach may assist in (partially) addressing an obstacle, an empty circle (○) that it may assist in assessing the severity of a problem, a check mark in a circle (⊘) that it may assist in both. RR: regulatory requirement. *Non-technical approaches*, technical approaches. "Distribution shift" and "unintended consequences" both refer to the deployment setting.

| | Solution approach | Training data scarcity | Distribution shift | Under-representation | Label errors | Data biases | Spurious correlations | Model underspecification | Insufficient o.o.d. tests | Unintended use | Automation complacency | Privacy concerns | Malicious attacks | ML system opacity | Unknown failure modes | Unfair discrimination | Unintended consequences | Comments and exemplary references |
|---|---|---|---|---|---|---|---|---|---|---|---|---|---|---|---|---|---|---|
| **Project inception** | *Risk assessment* | | | | | | | | | ○ | | | | | | ○ | ○ | RR. [80], [81] |
| | *Goal specification* | | ⊘ | ⊘ | | | | | | ⊘ | ⊘ | ⊘ | ⊘ | ⊘ | ○ | ⊘ | ⊘ | Incl. intended use (RR). [30], [81], [83] |
| | *Participatory design* | | ⊘ | ⊘ | ⊘ | ⊘ | | | | ⊘ | ⊘ | ⊘ | ⊘ | ⊘ | ○ | ⊘ | ⊘ | Involve stakeholders, especially most vulnerable groups, throughout. [30], [110] |
| **Data collection & preparation** | Targeted data collection | ✓ | ✓ | ✓ | ⊘ | ⊘ | ⊘ | ✓ | ✓ | | | | | | | ✓ | | Costly, cannot address all biases, privacy challenges. [111]–[122] |
| | Data augmentation, synthetic data | ✓ | ✓ | ✓ | ✓ | ✓ | ✓ | ✓ | | | | | ✓ | ✓ | | ✓ | | [116], [123]–[130] |
| | Over-/undersampling, weighting | | ✓ | ✓ | ✓ | ✓ | | | | | | | | | | ✓ | | [124], [131]–[134] |
| | Semi-supervised learning | ✓ | ✓ | ✓ | ✓ | ✓ | ✓ | ✓ | | | | | | | | ✓ | | [135]–[138] |
| | Federated, split learning | ✓ | ✓ | ✓ | | | | ✓ | ✓ | | | ✓ | | | | ✓ | | [139]–[146] |
| **Model selection & training** | Transfer learning | ✓ | ✓ | | | | | | | | | | | | | | | [135]–[137], [147]–[151] |
| | Model constraints, domain knowledge (incl. causal modeling) | ✓ | ✓ | ✓ | ✓ | ✓ | ✓ | ✓ | | | | | ✓ | ✓ | ✓ | | ✓ | Reduces degrees of freedom and required data set size. [152]–[158] |
| | Regularization | ✓ | ✓ | ✓ | ✓ | | | ✓ | | | | | | ✓ | | | | [159], [160] |
| | Online, continuous learning | ✓ | ✓ | ✓ | | | | | | | | | | | | | | Introduces new vulnerabilities, unclear how to guarantee safety. [161], [162] |
| | Privacy-preserving learning | ✓ | | ✓ | | | | | | | | ✓ | ✓ | | | | | [163]–[167] |
| | Adversarial training | ✓ | ✓ | | | ✓ | ✓ | ✓ | | | | | ✓ | | | | | [168], [169] |
| | Uncertainty quantification (UQ) | | ○ | | | | | | | | ✓ | | | | ✓ | | | [170]–[182] |
| | Interpretable modeling | | | | | ○ | ○ | | | | | | ✓ | ✓ | | | ○ | Requires domain-specific modeling, not always applicable. May detect attacks. [183]–[189] |
| | Post-hoc explainability methods | | | | | ○ | ○ | | | | | | ✓ | ✓ | | | ○ | Relatively easy to deploy, may be misleading. [183], [184], [190]–[196] |
| | Fairness-aware learning | | | | | | ✓ | | | | | | | | | ✓ | | Consider proxy variable choice [14], [197], [198], classifier splitting [199]–[203], causal learning [204]–[206], fair representation learning [207]–[209]. Constrained learning may be problematic [130], [210], especially in medicine. Standards [211], [212]. |
| **Model evaluation** | Evaluation on o.o.d. data (external validation) | | ○ | ○ | ○ | ○ | ○ | ○ | ✓ | | | | | | ✓ | | | Many current ML devices with regulatory approval do not have this. [38], [213], [214] |
| | Adversarial evaluation | ✓ | ✓ | ○ | | | ○ | | ✓ | | | | ○ | | ✓ | | | [215]–[219] |
| | Formal verification | ✓ | ✓ | | | | | | ✓ | | | | ✓ | | ✓ | | | [215], [216], [220]–[227] |
| | Penetration testing, red teaming | | ○ | ○ | | ○ | ○ | | ✓ | ○ | | | ○ | | ✓ | ○ | ○ | [228] |
| | Subgroup performance analyses | | | ○ | | | | | | | | | | | ✓ | ○ | | May be RR [40], [41]? Be aware of infra-marginality [229], [230] and potential biases [198], [231], [232], assess model performance and clinical impact [210], [232]–[234], consider causal analyses [206], [235]. Standards [211], [212]. |
| | UQ evaluation | | | | | | | | | | | | | | ✓ | | | [175], [176], [180] |
| | Evaluation of explanations | | | | | | | | | | | | | ○ | ✓ | | | [184], [191], [194], [236]–[238] |
| | *User studies* | | | | | | | | ✓ | ○ | ○ | | | | | ⊘ | ○ | RR. [83], [239] |
| | *Algorithmic impact assessment* | | | | | | | | | ○ | ○ | ○ | | ○ | | ○ | ○ | [30], [59], [65], [240] |
| **Model deployment & monitoring** | Distribution shift monitoring | | ○ | | | | | | | ○ | | | | | | | | [241]–[243] |
| | Out-of-distribution detection | | ✓ | | | | | | | ○ | | | ✓ | | ✓ | | | [176], [244]–[249] |
| | Uncertainty communication | | ✓ | | | | | | | | ✓ | | | | ✓ | | | [30], [245] |
| | Transparent reporting | | | | | | | | | | | | | ✓ | | ✓ | | [113], [250]–[254] |
| | *UI design* | | | | | | | | | ✓ | ✓ | | | | | | | [83], [255]–[259] |
| | *User training* | | | | | | | | | ✓ | ✓ | | ✓ | | ✓ | | | RR. [43], [83], [183], [238], [260], [261] |
| | *Procedural safeguards* | | | | | | | | | ✓ | ✓ | ✓ | | | ✓ | | | RR. [83], [262] |
| | *Post-market surveillance* | | ○ | | | | | | | ○ | | | | | | | ○ | RR. [78], [81], [87] |

The following section discusses the most critical barriers to achieving safety, robustness, and reliability in medical machine learning systems. Section III-B then describes technical measures to increase model robustness and reliability, which can both be seen as necessary requirements to *prevent* the occurrence of failures. Finally, section III-C discusses methods for fault *detection* and *mitigation*.

### A. OBSTACLES TO ACHIEVING SAFETY, ROBUSTNESS, AND RELIABILITY

Possibly the most salient challenge to overcome for achieving safe, robust, and reliable MML concerns the lack of large, representative, high-quality datasets. The high dimensionality of the input space in most application scenarios means that it is usually impossible to sample the input space sufficiently densely without relying on vast amounts of training data. Moreover, training data collection may be limited to specific countries, hospitals, medical devices, participant acquisition channels, or further circumstances that may constrain the types of data observable *in principle* using a specific data collection process. Consequently, models are often used to make "out of distribution" (o.o.d.) predictions, i.e., extrapolate to previously unseen regions of the data space. Unfortunately, existing medical databases are typically confined to small, inaccessible data *silos* [272], and riddled with labeling errors [273]–[275] and biases [276]–[281]. There are various valid reasons for this, including legal constraints,[6] concerns over patient privacy, the low prevalence of rare diseases, and the high cost of performing studies [282]. Data labeling, segmentation, and annotation are major sources of errors and require extensive effort, often by medical experts [272]. Inter-observer variability, the inherent uncertainty in the medical domain, and plain labeling errors all contribute to label noise [274]. Moreover, since an exact and accurate ground truth label is often unavailable, data labeling often depends on interpretations of feature relevance and labeling styles [272]. *Hidden stratification*, i.e., summarizing multiple distinct groups, such as different sub types of lung cancer, under a single label, can lead to large differences in performance across these subgroups [283], underlining the importance of selecting an accurate and fine-grained labeling scheme. For all of these reasons, the gathering of large, high-quality, representative datasets is widely recognized as a decisive challenge for the successful application of machine learning in the medical context [1], [282], [284]–[286].

On a more conceptual level, the traditional statistical learning framework assumes that the data used for training the algorithm and the real-world target data to which the classifier will ultimately be applied are drawn from the same data generating distribution. This assumption is rarely met in practice [287], [288]: usually, data observed "in the wild" (the *target domain*) differ in many characteristics from those in the training data[7]; a phenomenon called *distribution shift* or *dataset shift* [287], [288]. Distribution shift may occur for many different reasons, including sampling biases and different instrumental or environmental noise in the target data and the training data. In the context of medical image analysis, potential causes include differing viewing angles, movement of the captured objects, lighting conditions, or different optical sensors [22]. In particular, it is a well-known challenge to train ML models that work across different scanners or imaging protocols [289]. A robust ML model would overcome such rather superficial distribution shifts and would perform as expected. Another type of distribution shift may arise when the training data are *inherently* different from the target data. Using images of parts of dead bodies to train a model to be applied to living patients is one example from the medical domain; the use of electronic health records gathered from young, healthy persons to train a model that will (also) be applied to sick and older patients is another. In these cases, again, the essential information required for solving the task is assumed to be present in the training data nevertheless, but a non-robust machine learning model may fail to distinguish relevant and irrelevant features and may thus perform poorly on the target data. Unfortunately, modern deep learning methods such as those used in medical imaging analyses are particularly brittle with respect to even minimal dataset shifts [290]–[293] — in other words, they are not robust. In the medical domain, dataset shift has thus been identified as a critical challenge preventing the widespread application of MML methods because, e.g., trained models are not robust against differing recording environments or devices, patient demographics, hospital types, healthcare systems, or healthcare policy shifts [11], [138], [289], [294]–[296]. Notice, however, that not all dataset shifts necessarily pose a problem, and some may even be beneficial: intentionally over-representing minority groups in the train-ing data is one example that will be discussed in detail further below.

---

[6]Important regulations in this regard include the general data privacy regulation (GDPR) in Europe and the health insurance portability and accountability act (HIPAA) in the US.

[7]By *training data*, we refer to the union of the training, validation, and test datasets. If it is necessary to refer to the first of those three exclusively, we will refer to it as the training data*set*. This distinction will rarely be necessary, however.

*Spurious correlations* represent a related yet different challenge: it is well known that irrelevant features often correlate with the prediction label [116]. These spurious correlations are usually not a problem when using a white-box model structure that exploits the available prior knowledge about relevant and irrelevant factors. Highly flexible deep learning models, however, incorporate very little — if any — prior knowledge, and are thus prone to exploit these spurious correlations for improving their predictive accuracy on the training data; a phenomenon often called *shortcut learning* [116]. Examples of confounding factors that have been found to (inadvertently) strongly influence model predictions include patient and hospital process data [10], surgical skin markings [12], hospital system and department within a hospital [11] and various other factors including age and gender [297]. Recently, Zhang *et al.* [298] have shown that adversarial examples can be understood as intentional exploits of spurious correlations learned by the model.

Both their susceptibility to spurious correlations and their non-robustness to distribution shift are consequences of modern deep neural network (DNN) model's vast amount of parameters and neglect of any prior knowledge. These features grant them the expressiveness to capture every statistical peculiarity present in the training data. Moreover, modern DNNs are sufficiently expressive for many different models to achieve very similar predictive performance on data drawn from the same distribution as the training data; a situation called *model underspecification* [299]. D'Amour *et al.* [299] show that this situation is omnipresent in modern (deep) ML pipelines and that it results in highly variable real-world performance: a subset of the large class of models achieving similar performance on the training distribution may perform well on (inevitably slightly different) real-world data [116], but whether a model from this subset is actually learned — and not one that transfers badly to the real world — is more or less random [299]. Thus, in addition to the trained models, the *training process* is typically also non-robust: the resulting models may differ widely if supposedly unimportant details such as the random seed, the initialization method, or the particular implementation of the algorithm are changed [299]. Figure 3 illustrates the relationships between dataset shift, shortcut learning, and model underspecification.

Finally, aggravating the aforementioned challenges, the performance of ML systems is often evaluated solely using i.i.d. data [214], thus *not* testing their (generalization) performance on o.o.d. data. This has been argued to be a crucial cause of model brittleness not being discovered during system development [4], [213], [254], [299], especially with modern ML methods that have the capacity for strong overfitting.

### B. MEASURES TO ACHIEVE ROBUSTNESS AND RELIABILITY

In the following, we will discuss various technical measures for achieving more robust and reliable MML systems,

proceeding stage by stage through the lifecycle of an MML application.

#### 1) PROBLEM SPECIFICATION

At the beginning of the problem specification phase, the target distribution should be carefully described. This distribution depends upon many factors, including the clinical use case, the patient population, target measurement devices, and recording environments. This first step is crucial for selecting appropriate robustification strategies against expected distribution shifts and for later identifying an expected or unexpected distribution shift during deployment. If it is difficult to obtain a model that is robust and reliable across the originally intended target domain, it may be necessary to restrict the target domain. This may mean limiting the application to be used within a particular patient group,[8] only with a small, specific set of sensors, or with a certain camera angle and lighting, to give some examples. Moreover, it needs to be determined whether the introduction of adversarial examples by malicious attackers is of concern; refer to the following section IV for a discussion of adversarial robustness. Besides its technical necessity, a precise specification of the MML system's *intended use* is demanded by existing regulations.

#### 2) DATA COLLECTION AND PREPARATION

It is well recognized in the ML community that the availability of a large, representative and diverse dataset is an essential ingredient to train a robust and reliable ML model [112], [113]. As a first and foremost requirement, a dataset should contain sufficiently many examples from each relevant target group. In practice, this is usually not realistic to achieve due to the high dimensionality of the input space in most application scenarios, as was discussed in :reliability-obstacles. Thus, as out of distribution (o.o.d.) prediction is generally unavoidable, testing the model's performance in such o.o.d. prediction settings (in other words, its generalization capability) is crucial [38]. One strategy to assess a model's generalization capabilities is to intentionally introduce o.o.d. examples in the *test* dataset, i.e., examples that differ in some characteristic from the examples included in the training dataset (cf. fig. 3). In practice, training, validation, and test datasets are often carefully (and iteratively) crafted to include a sufficient amount of examples from all regions of the target data distribution [112]. To achieve this, it is often useful to significantly over-represent minority groups, e.g., patients with rare diseases or ethnic minorities, in the training data (as compared to the target data distribution), because otherwise such groups would only be sparsely represented, and the resulting model might perform poorly on such rare groups [114], [287]. In this sense, it may be helpful to use a training distribution differing widely from the target distribution: as an example, it is often beneficial to employ a

---

[8]Of course, applicability to a large and diverse group of patients should always be the goal. Developing products only for certain patient groups can be seen as unfairly discriminating against other groups, depending on the circumstances and reasons for doing so.

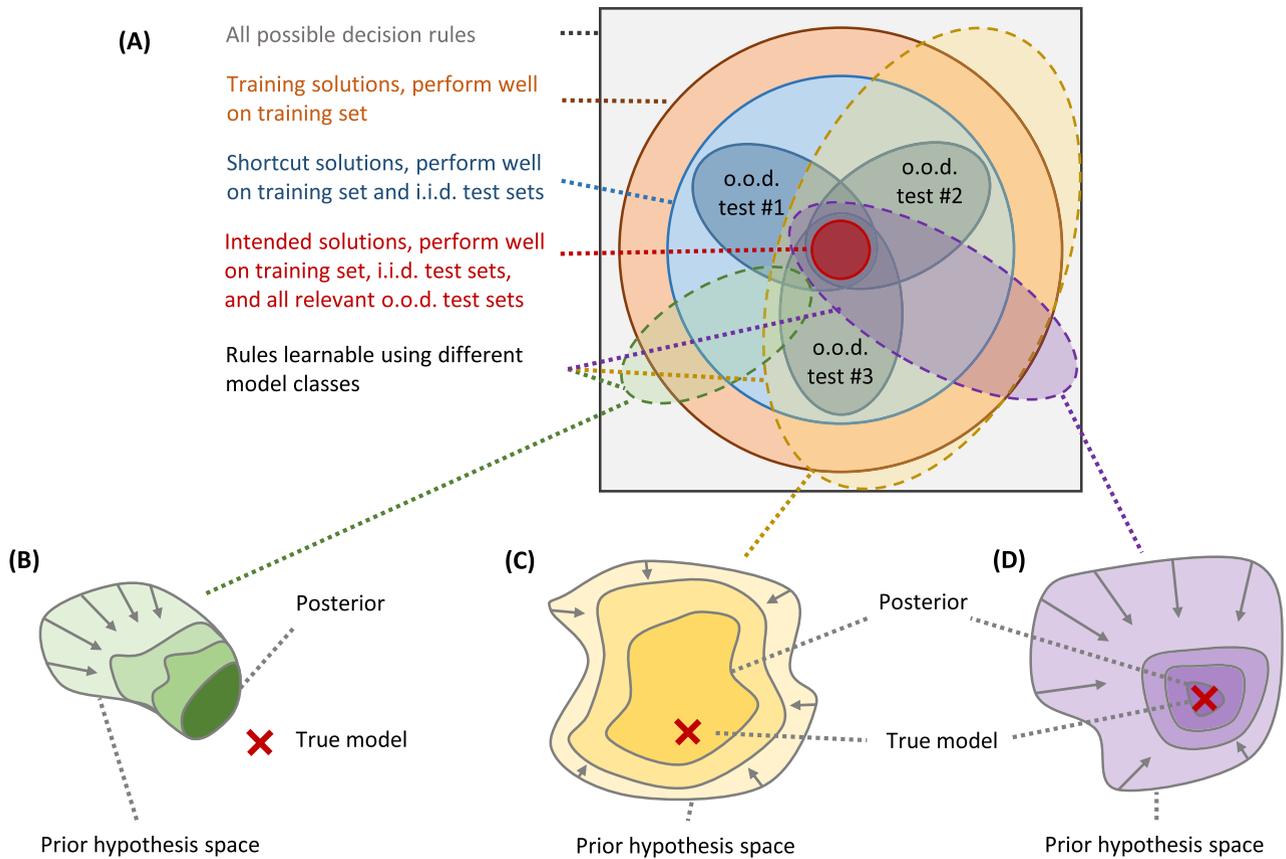

**FIGURE 3.** (A) Models that perform well on the training dataset need not perform well on the test dataset, and those performing well on the test dataset still need not perform well in the real world, for reasons including dataset shift, spurious correlations, and model underspecification. Figure closely modeled after Geirhos *et al.* [116]. Independent identically distributed (i.i.d.), out of distribution (o.o.d.). (B) A model class with a too restrictive hypothesis space prevents convergence towards the true model. (C) A model class with inadequate inductive biases renders convergence towards the true model inefficient, requiring lots of data. (D) A model class must cover a sufficiently large hypothesis space that includes the true model, it must be equipped with adequate inductive biases *and* there must be sufficient data to enable convergence towards the true model. Figures (B)–(D) closely modeled after Wilson and Izmailov [300].

uniform training distribution in instances where the target distribution is strongly imbalanced [114], [132], [301]. Such a reweighting then corresponds to over-representing minority groups and under-representing majority groups in the training dataset. Dataset weighting methods such as propensity score weighting [134] or importance weighting [131] are often employed to implement such a re-balancing. Aside from dataset balancing strategies, the intentional introduction of training examples that contradict potential spurious correlations represents another valuable dataset curation strategy that may help prevent shortcut learning [116]. Notice that such dataset curation strategies actively induce data shifts — which are, however, hoped to exert a beneficial effect on the training process and outcome — and also compare section VI for a discussion of the effect of dataset curation on the fairness of the resulting model. In this regard, it is also worth mentioning the "All of Us" initiative, which "plans to enroll a diverse group of at least 1 million persons in the United States" [111], with most of the participants stemming from groups that were previously underrepresented in biomedical datasets. Similarly, the proposed "European

health data space" may represent an essential step in the right direction if executed well.[9]

To alleviate the (difficult to fulfill) need for gathering large-scale, application-specific, centralized databases containing sensitive patient health data, distributed learning methods such as *federated* learning [139], [140] and *split* learning [302] have been proposed. These methods allow training a model using data from different sites without requiring them to share their sensitive data with any other site or a central entity. This is achieved by running partial model update steps locally at each site and then sharing the result of these computations (but not the data they are based on) with other sites or a central entity. Federated learning and split learning are closely related and differ mainly in the way the partial model computations are split across the participants. Each method has been found to be preferable performance-wise under different circumstances [144], and hybrid methods have been proposed to combine the benefits of both [143], [303]. Gao *et al.* [144] provide a comprehensive performance evaluation of both approaches for varying types of datasets, models, and (computing or communication) performance constraints. Due to the particular challenges related to the gathering of large medical datasets, distributed learning has been prominently put forth as an essential ingredient for solving the data silo problem in medical machine learning [101]–[103], [141], [142]. Various distributed learning toolboxes have been developed [304]–[306], and the IEEE has recently published a guidance document on the development of federated learning frameworks and applications [307]. Owing to their distributed nature, however, these methods raise new challenges regarding privacy and security, which will be discussed in detail in section IV.

---

[9]See https://ec.europa.eu/health/ehealth/dataspace_en

*Semi-supervised learning* represents an alternative avenue for reducing the effort necessary to create large datasets by leveraging unlabeled data in addition to a smaller set of labeled data [135]. Gathering sufficiently large labeled datasets requires an amount of manual labeling effort by medical experts that may simply not be admissible in many cases. As an example, Bai *et al.* [136] propose a semi-supervised learning method for cardiac MR image segmentation. Semi-supervised learning methods depend on some basic assumption about the underlying data distribution, such as smoothness of the response surface (similar input data have similar labels), that decision boundaries pass through low-density regions of the data space, or that data sharing a label lie on low-dimensional manifolds [135]. The utility of using additional *unlabeled* images in the context of medical imaging has recently been questioned [138] because additional unlabeled examples can only provide further information about the input data distribution but *not* about the mapping between input data and their target labels. A slightly different approach to solving the labeling problem is *automatic labeling*. These methods typically exploit more domain-specific knowledge than general semi-supervised learning methods. Trivedi *et al.* [137] have proposed a machine learning model for automatically extracting mammogram labels from their accompanying free-text clinical records. The authors found their method to perform well in the case of short text records. Similar methods have been proposed for extracting information on pulmonary emboli from free-text radiological clinical records [308]. In a slightly different vein, Yi *et al.* [309] trained a deep neural network to label chest mammograms semantically with their mammographic view and breast laterality.

*Transfer learning* in general and *domain adaptation* (which is a sub-field of transfer learning) in particular represent other alternative approaches for increasing the amount and representativeness of the available training data. The fundamental idea is to employ available data from different (but in some sense similar) domains for improving model performance. There are various approaches within this field, from using models pre-trained on general-purpose labeled image databases and specializing them to a medical image recognition problem [147] to using data recorded from different patient groups or using different recording devices.[10] One instance of the latter approach is called *domain representation learning*; its aim is to identify a set of feature transformations that accurately represents data recorded from different domains, e.g., histological images from different labs [148].[11] The actual classification or regression model is then trained on the data in this identified domain representation, thus making it applicable to data from many different domains and enabling the use of data from superficially different domains in the training phase. Notice that standard preprocessing techniques like normalization and motion correction [311] can be understood as enforcing a particular kind of domain representation. For recent reviews and applications to biomedical imaging, refer to, e.g., [149]–[151].

If a sufficiently large and representative dataset is unavailable, (partially) *synthetic data* may be used to augment or completely replace real data [123]. Many methods fall into this broad category, from image data augmentation [138], [312] and synthetic resampling methods [124], [132], [132], [301], [313] to (patho)physiological simulation models [125], [126] and GAN-based synthetic medical data generation methods [314], [315]. As an example from general-purpose image recognition, Geirhos *et al.* [316] demonstrate that training a model on a *stylized* version of the ImageNet database, which equips objects with unusual textures, biases the trained model towards recognizing shapes instead of textures, thereby significantly increasing robustness. When using synthetic data, a crucial question regards the proper choice of the model parameters for generating a synthetic dataset that optimally supports learning a robust and reliable model. *Domain randomization* is one approach that has been proposed for solving this problem by incentivizing successful concept learning [127], [128]. However, many questions remain unanswered: what are the requirements on the simulator and the parameter selection method for training a robust and reliable model using synthetic data? Should one use only synthetic data or mixed datasets? Moreover, just like above: how does one synthesize a *balanced* dataset? Recent evidence against the utility of synthetic dataset augmentation has been provided by Taori *et al.* [293], who found that adding various types of synthetic distribution shift — e.g., artificial noise or image rotations — did *not* increase the resulting model's robustness against naturally occurring distribution shift. This observation is in contrast to the results of Michaelis *et al.* [317], who found simple training image stylizing [129] to significantly increase model robustness. Many groups are currently investigating the use of synthetic data for machine learning; thus, it appears likely that answers will emerge soon. For a recent review regarding the use of synthetic data in deep learning, refer to Nikolenko [123] and Fawzi *et al.* [312].

---

[10]As was mentioned briefly in section II-A, there are complex regulatory questions associated with the use of transfer learning [90].

[11]Specific representation learning methods have also been advocated as tools to increase the interpretability [310] or fairness of the resulting MML model, cf. section VI for details.

A final aspect regarding the employed datasets concerns their preprocessing. Application-specific preprocessing steps that exploit relevant domain knowledge may significantly simplify the subsequent estimation problem. As one recent example, Li *et al.* [318] demonstrate improved chest X-ray classification accuracy when using a specific preprocessing step that suppresses bones in the recorded images. Concerning the handling of errors in the employed datasets, one needs to distinguish between different types of errors. If *measurement outliers* frequently occur in the target data, then they should also be represented in the training data — in this way, a prediction system (potentially including a preprocessing step that discards such outliers) can be designed that is robust to these outliers. *Labeling errors* (or, in the case of regression, outliers in the target signal), on the other hand, should be avoided in the training data, wherever possible. Note, however, that a *robust* learning method — as discussed above — should yield a similar model even in the presence of a few labeling errors. It has been claimed in the literature that deep learning models are inherently robust to label noise in the training dataset [319]. In light of the recent analyses of D'Amour *et al.* [299], which question the validity of test set performance (for underspecified models) as an indicator of real-world performance, it is unclear, however, whether these results translate into practice. In a similar vein, Northcutt *et al.* [320] recently reported that labeling errors in the *test* dataset (as opposed to the *training* dataset, as analyzed by, e.g., Rolnick *et al.* [319]) crucially affect real-world performance because they influence model selection. The obvious countermeasure against label errors in training and test datasets is an increased investment in the collection of high-quality data. More generally, it is crucial to perform an in-depth data quality assessment, for which various data quality indicators have been proposed and (health data specific) toolboxes are available, cf., e.g., Schmidt *et al.* [117]. Efforts for the standardization of data quality assessments are underway, both medicine-specific [107] and concerning general ML [118].

### 3) MODEL SELECTION AND TRAINING

The canonical method to achieve robustness of the training process is to take some kind of prior knowledge about the class of reasonable models into account. This will bias the training process towards models that concur with these prior assumptions. In the simplest case, one can incorporate prior knowledge by assuming a simple, parametric model structure. While the use of simple model structures may, in some cases, increase model robustness, this may come at the expense of overall prediction performance. Notably, a recent systematic review of clinical prediction models in a large number of studies found no evidence of superior performance of ML methods over logistic regression [4]. Similarly, Jacobucci *et al.* [8] find no benefit to ML methods over linear models for clinical outcome prediction after accounting for improper performance evaluation methodology. However, prior knowledge can be injected into the estimation process in a myriad of different ways, including choosing a specific (grey box) model structure, the specification of Bayesian priors over various model parameters, or imposing hard constraints on model parameters and properties.[12] One prominent line of research in this field concerns the training of models (including, in particular, deep neural networks) that satisfy some prediction monotonicity constraint with respect to one or multiple input characteristics [152], [153]. Such a monotonicity constraint, which represents a reasonable yet very mild assumption in many application scenarios, may already significantly improve model robustness. Similarly, network structures have been proposed that exploit symmetries and invariance properties, such as invariances with respect to image rotation, translation, and reflection [154], and Evans and Grefenstette [321] propose a method to combine the flexibility of neural networks with the data efficiency of inductive logic programming (ILP). In another branch of research, Barnett *et al.* [159] and Chen *et al.* [155] have proposed slightly modified deep learning architectures that utilize a special *prototype layer* which characterizes the similarity of the input data with prototypical training dataset examples. While this modification was mostly introduced to increase the interpretability of the model (cf. section V), this also represents an inductive bias that may reduce the likelihood of learning spurious correlations during training. (Notice, however, that such methods are highly sensitive to the choice of the employed prototypes or concepts.) Importantly, as one would expect, recent research demonstrates that the incorporation of such mild constraints on the learned models (if they are well-justified) does *not* diminish predictive model performance [153], [155], [159]. Such constraints may, on the contrary, help alleviate the underspecification problem by restricting the feasible model space [299] (also see Figure 3).

Another branch of research investigates causality-aware models that impose constraints on the relationships between causes and effects, and thereby increasing robustness to dataset bias, distribution shift, spurious correlations, and adversarial examples [156], [295], [298], [322]–[324]. It has been argued that classical ML models' inherent unawareness of the possible interactions of causes and effects imposes a fundamental limitation to the level of performance they can achieve [156], [323]. Shimoni *et al.* [325] have developed a toolbox that implements causal inference methods and the evaluation schemes required for their practical application in the medical domain.[13] Also in the medical domain, Richens *et al.* [157] have recently shown causal inference to significantly outperform purely associative inference, as performed by usual ML methods, on a differential diagnosis task. Subbaswamy and Saria [295] have discussed the merits of using causal models in healthcare for achieving robustness to distribution shift. Similarly, Castro *et al.* [138], [326] have argued that a causal perspective is crucial for machine learning for medical imaging and have provided a framework for doing so. Moreover, Holzinger *et al.* [158] have argued that causality is a necessary prerequisite to achieve real explainability, which will be the subject of section V. section VI will discuss the beneficial properties of causal models for achieving algorithmic fairness.

---

[12]Such hard constraints can, of course, also be formulated as probabilistic priors.
[13]See https://github.com/IBM/causallib

Especially in medical applications, precisely quantifying the model's uncertainty in making a given decision is essential [175], [327], [328]. Proper uncertainty quantification is a challenging task — in particular on o.o.d. data [174] — and requires a careful analysis and characterization of general model reliability. While standard, unconstrained cross-entropy minimization tends to produce classifiers that are reasonably well-calibrated on in-domain data [329] and can be fine-calibrated with relatively straightforward post-hoc methods [171], this is generally not the case for o.o.d. data. Standard neural networks, in particular, are known to suffer from *asymptotic overconfidence*, becoming increasingly confident in their predictions the further away a sample lies from the training data distribution [174]. This represents an important challenge in particular in the healthcare context, with only a limited amount of data available, heterogeneity in patient characteristics and disease prevalence between sites and settings, and distribution shift over time [175]. There are both Bayesian [173], [178], [179] and non-Bayesian methods [170], [172], [177] for o.o.d. uncertainty quantification, some of them relying on o.o.d. data being available for calibration [177]–[179]. Many groups have used dropout as a cheap method for uncertainty quantification [170], including in the medical domain [182], but it has recently been shown that dropout represents a poor approximation to the true posterior [181]. Notice also that there is a close relationship between o.o.d. uncertainty quantification and out-of-distribution detection methods [176], [244]–[249]. For a comprehensive review and introduction to different techniques for uncertainty quantification in deep neural networks, refer to Gawlikowski *et al.* [180].

Finally, classical regularization techniques, such as weight decay or smoothness regularization, also increase model robustness by incorporating some prior knowledge about the system [160], thereby reducing the chance of overfitting the available data.

### 4) MODEL EVALUATION

In many practical ML applications, it is current practice to iteratively refine the test and training datasets based on field failures: it is observed that the current model performs poorly on some groups; thus, more examples of that group are added to the training and test datasets. In this way, the representativeness of these datasets is iteratively improved long after the initial deployment. While this may be a good practice in a low-stakes environment such as online advertising, it is, of course, limited in its applicability to medical ML: in the healthcare setting (as in many other physical settings), developers cannot afford to deploy poorly working prototypes and learn from their mistakes; the first deployed version of an MML system must *already* perform safely, robustly, and reliably [330]. Thus, utmost care must be taken to ensure the comprehensiveness of the model evaluation process.

Just like a large and representative *training* dataset is crucial for training a robust and reliable model, a large and representative *test* dataset is crucial to enable an accurate assessment of the trained model's performance in the real world. To analyze whether model performance suffers from hidden stratification, i.e., whether performance differs significantly between (unknown) subgroups of labeled groups, unsupervised learning techniques such as simple clustering algorithms can be employed [283]: if the model performs poorly on some of the identified clusters, these might be worthy of further investigation. To assess a model's generalization capabilities, it is essential that at least part of the test set represents out of distribution (o.o.d.) data, i.e., data drawn from a distribution that differs from the training distribution [213], [254], [299], [331].[14] (Also compare fig. 3.) These can be, e.g., from a different study, different hospital, different population group, different recording time, or obtained using different measurement equipment [213], [254], [331]. Unfortunately, many of the ML-based devices currently approved by the FDA have only been evaluated at a single site [38]. Using proper statistical evaluation methodology is, of course, essential, as improper evaluation methods have resulted in significantly exaggerated performance claims in various medical domains in the past [4], [8], [332].

In addition to data-based validation, various formal verification methods have been proposed for machine learning models, akin to classical software verification techniques. Many of these methods can be understood as specific forms of automated o.o.d. testing, as they automatically generate large numbers of test cases that (usually) deviate in some way from the examples contained in the training set. One benefit of formal verification methods is that they can enable assessing compliance with lawful requirements (e.g., nondiscrimination), as may be necessary to enable users to *contest* a decision made by the system [92]. While computational complexity is high, novel efficient relaxation techniques have been proposed to enable the application of Satisfiability Modulo Theory (SMT) solvers to deep learning models [220], [221]. These solvers can be employed to formally verify various properties of the input-output behavior of a learned model. Katz *et al.* [220] employed their proposed techniques to verify application-specific desirable properties of a real-world airborne collision avoidance system for unmanned aircraft. In the medical domain, Guidotti *et al.* [222] have employed an SMT-based technique for formally verifying a desirable input-output property of a prosthesis myocontroller. Because such SMT-based techniques are computationally very demanding and not applicable to arbitrarily large networks, Pei *et al.* [223] have proposed an alternative method (VeriVis) using input space reduction techniques. They utilize this method to verify multiple practically relevant safety properties of the input-output behavior of (then) state-of-the-art computer vision systems, thereby discovering thousands of violations of these safety constraints by various competitive academic and commercially available vision systems. These detected safety violations can then of course be incorporated into the training dataset, thereby hopefully increasing the robustness of the learned model. Michaelis *et al.* [317] discuss a range of additional tools that serve similar verification purposes for deep learning models. Of course, similar formal verification techniques have also been proposed for other classes of machine learning models, including, in particular, various kinds of tree ensembles [224]–[226]. Several of these formal verification methods can also be used to iteratively "repair" a model that does not (yet) satisfy the specified constraints until it does [222]. Finally, there is a large body of work regarding the formal verification of (e.g. $\ell_\infty$) adversarial robustness, which — as mentioned above — is closely related to the model's smoothness [215], [216], [220], [227]. In particular, the field of *semantic adversarial deep learning* is concerned with verifying the robustness of the learned model to semantic variations, i.e., variations that are of application-specific interest as opposed to purely synthetic noise [217]. Similar to the verification techniques discussed above, such semantic adversarial examples that have been found to be violated by a trained model can then be included in the training dataset to achieve model robustness towards this type of variation.

---

[14]In some fields, this is called *external* or *internal–external* validation [213], [254], [331].

As a last remark on model evaluation, interpretable models are obviously easier to evaluate thoroughly than black-box algorithms since the developer and other stakeholders can (manually or in an automated fashion) assess the model's reasoning behind a particular decision, thereby increasing the likelihood of detecting erroneous behavior before model deployment [29], [333]. This represents one of the principal arguments for employing interpretable models, which will be the subject of section V.

### 5) MODEL DEPLOYMENT AND MONITORING

During the final stage of an MML model's lifecycle, continuous monitoring for possible distribution shift is a permanent and essential task. Is the model being employed in an environment it was not trained for, or using sensors that have not been part of the training data, or on unforeseen patient groups? Besides gathering general statistics about the target data distribution, one may want to regularly draw random samples from the target distribution and re-evaluate the model accordingly. Aslansefat *et al.* [241] have recently proposed a formal method for monitoring and quantifying distribution shift, potentially serving to recognize when a model is used outside of its safe application area. To detect and quantify *label shift* specifically, Lipton *et al.* [243] have proposed Black Box Shift Estimation (BBSE). In practice, however, simple statistical metrics are often used to monitor dataset shift [242]. Crucially, feedback from the end-users — patients or healthcare workers — regarding the performance of the model and its potential failures should also be gathered and incorporated continuously. While distribution shift monitoring is not explicitly required, current regulations do demand extensive post-market surveillance activities for medical devices (cf. the discussion in section II-A).

### C. MEASURES TO ACHIEVE FUNCTIONAL SAFETY

Model robustness and reliability are necessary but certainly not sufficient requirements to achieve an MML application's functional safety. These two properties both serve to *prevent* the occurrence of failures. There are, however, more strategies that can be used to this end. For one, *transparency* regarding safe operating conditions is crucial: Under which circumstances and on which dataset was the model trained? Which are the assumed conditions on the operating environment and the patient characteristics? These requirements must be clearly stated and communicated to the user, and deviations from these required conditions should be detected automatically where possible, requiring manual human decision-making in these cases. Similarly, automated rejection techniques which aim to recognize situations in which the model's predictions are unreliable because it is performing "out-of-distribution prediction" can be used [176], [244]–[249]. Besides being transparent about the required operating environment, another important systematic risk mitigation strategy is to *ensure* that these operating conditions are met by means of *procedural safeguards* [262] such as, e.g., installing a sensor at a fixed position, designing a system that incorporates an ideal lighting solution, only allowing sensors by particular manufacturers, or designing a user interface that requires the doctor to check that the patient belongs to a supported patient group. On an organizational level, akin to classical software security measures, *red teaming*[15] and *bias and safety bounties* have recently been proposed as effective methods for increasing the safety of ML systems [228]. (Notice that these, again, can be understood as measures to achieve comprehensive o.o.d. model evaluation.) Finally, proper operator training is essential [43], [260], [261]: users — be it patients or healthcare workers — must be educated about the capabilities and limitations of a machine learning system in order to prevent incorrect and potentially unsafe usage. Safety measures such as the ones mentioned in this paragraph fall under the umbrella term of *usability engineering* (as described, e.g., by IEC 62366-1 [83]) and are an essential requirement for regulatory conform MML.

If an error occurs despite all countermeasures, how can its negative impact be mitigated? Firstly, the error should be *detected*. Automated rejection techniques, which detect inputs for which the model is particularly uncertain, or for which only a few similar examples exist in the training data, have already been mentioned above. On the human side, again, interpretability of the model (cf. section V) is highly beneficial: if the user can understand the model's reasoning behind a decision, the assumptions that went into it, as well as its uncertainty, this may serve to recognize faults that might otherwise remain unnoticed. To leverage this human capability to detect system faults, preventing *automation bias* and *automation complacency* is crucial [161], [255], [256]. It has been demonstrated [255], [256] that in many practical scenarios, human operators — especially when under time pressure and multi-tasking — quickly place too much trust in an automated system which has performed reliably in a large number of cases, thereby increasing the likelihood of a system fault not being detected by the operator. In the medical domain, Gaube *et al.* [257] have recently found that physicians with little task expertise tend to trust advice they receive on a diagnostic task, regardless of whether it is from an AI system or a colleague. Proper user training and emphasizing the user's responsibility have been demonstrated to reduce the prevalence of automation bias in healthcare workers operating clinical decision-making systems, alongside careful user interface design and reporting of confidence ratings [256]. Both proper user training and deliberate user interface design are key elements of medical device development processes [83] and should, of course, be evaluated in user studies [239].

---

[15]See, e.g., https://www.wired.com/story/facebooks-red-team-hacks-ai-programs/

Finally, as a last resort, medical ML systems should react robustly to any failures that might occur despite all prevention measures, i.e., perform failure *mitigation*. This becomes especially important once the ML system automatically interacts with the patient in a closed loop.

### D. SUMMARY: SAFETY, ROBUSTNESS, AND RELIABILITY

To achieve functional safety of ML-based medical systems, failures must be *prevented* where possible, *detected* if they occur nevertheless, and *mitigated* to ensure that the system remains in a safe state at all times. In MML systems, model robustness and reliability represent essential requirements for failure prevention. Unfortunately, these properties are inherently difficult to assess with regards to neural network models because good test set accuracy often does not translate into good real-world performance due to distribution shift [287], [293], model underspecification [299], test set label errors [320], and purely i.i.d. testing, among other challenges. The use of large, representative datasets (which may be aided by employing federated learning methods, synthetic data augmentation, or automated labeling schemes), the incorporation of prior knowledge about the class of reasonable models, and extensive out of distribution testing are key ingredients to achieve models that are robust and reliable. However, to ensure the safety of an MML system, one should not stop at technical considerations about the model but also consider measures to increase the safety of the whole system, such as operator training, proper user interface design, and physical enforcement or automatic verification of safe operating conditions. Such *usability engineering* methods are required by current medical device regulations as a risk mitigation strategy, and MML developers can draw on decades of experiences and standards in this field [83]. However, important transfer work in developing functional safety concepts (and standards) for MML applications remains to be done [264]. A unique role is played by transparency — regarding, e.g., the training data composition, model structure, training process, and final trained model — which increases robustness and reliability throughout all lifecycle stages of the MML system, and which will be the exclusive subject of section V. Finally, section VI will discuss the problem of fairness and nondiscrimination, which is closely related to reliability: if a model performs equally well across all relevant patient groups, it is both reliable and fair (at least concerning this particular definition of fairness).

## IV. PRIVACY AND SECURITY

Medical machine learning applications demand an exceptionally high degree of data privacy and application security. Beyond classical privacy and security measures, this necessitates guaranteeing the privacy and security of the employed ML methods. While providing (possibly restricted) public access to the MML system is unavoidable if it is to be used in practice — users must at least be able to provide inputs to the system and receive model predictions in return — it should at no point during the system's lifecycle be possible for an outsider to extract sensitive patient data. Moreover, malicious actors must be prevented from negatively influencing the outcome of the training process or a particular prediction. Spurred by the urgent practical and theoretical challenges this poses to the realization of privacy-preserving and secure machine learning processes, recent years have seen a proliferation of research on subjects such as privacy-preserving ML and federated learning. In this context, an important role is played by different communication schemes that can be pursued, and that determine who has access to the training data, the trained model, and a patient's query data that serve as inputs to the trained model. Figure 4 depicts four commonly used communication schemes that may be used in practice. The following sections will discuss the vulnerabilities of each of these setups, as well as possible defenses against malicious attackers.

### A. OBSTACLES TO ACHIEVING PRIVACY AND SECURITY

The challenges of achieving privacy and security in ML healthcare applications are (at least) sixfold:

1) classical cybersecurity,
2) confidentiality of the training data (models shall protect their training data),
3) confidentiality of the query data (patients that use a model shall not expose their patient data),
4) confidentiality of the model (preventing the extraction of the model from a cloud service),
5) integrity of the training process (malicious actors must not be able to influence the final model negatively or extract training data), and
6) integrity and robustness of the model (adversarially crafted inputs shall not confuse the model).

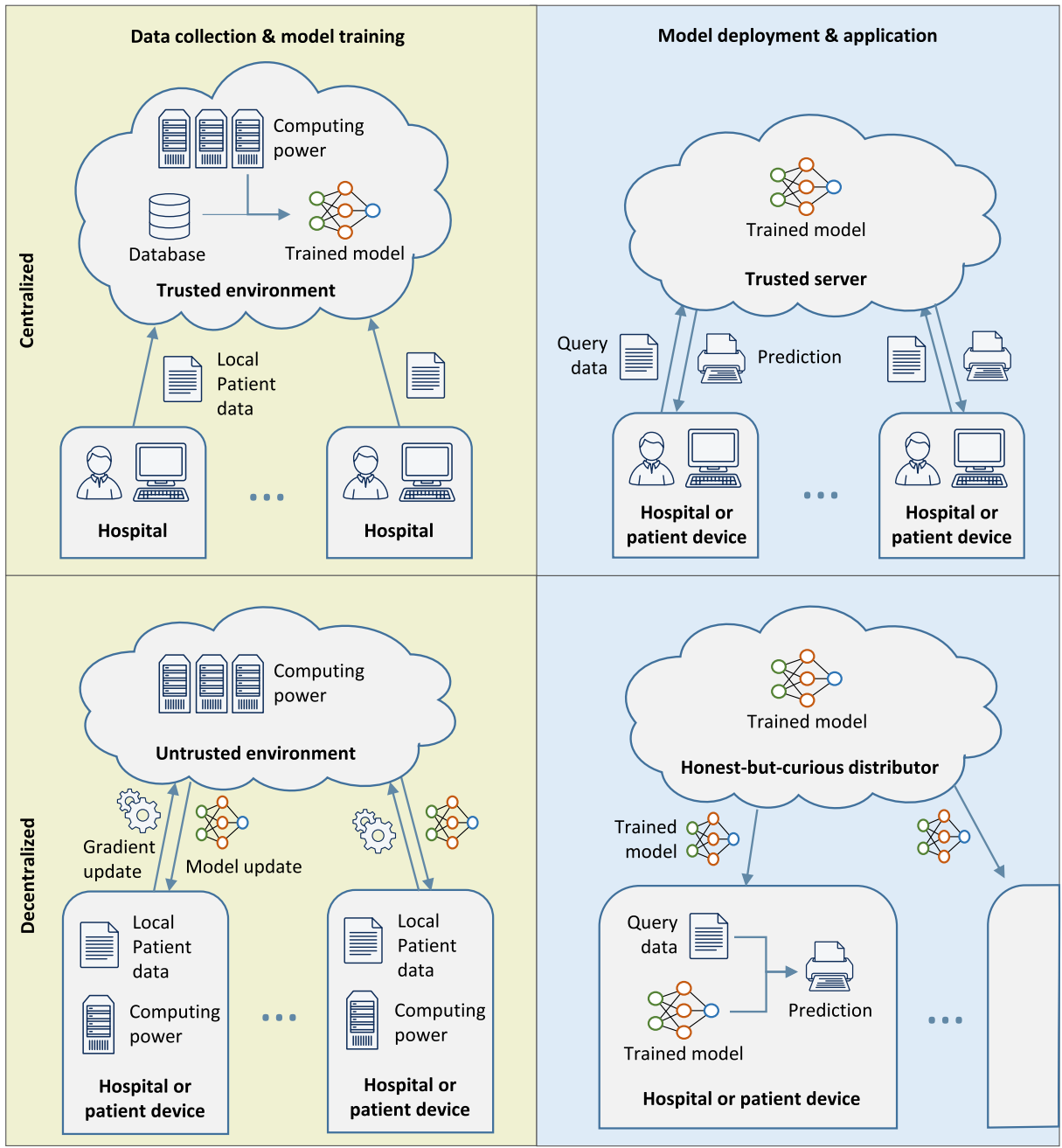

**FIGURE 4.** Many different communication schemes have been proposed for data collection, model training, and model deployment. All of them have merits and drawbacks—many of which will be discussed in section IV-A—and are preferable under different circumstances. Mechanisms for preserving privacy and security in these different schemes against various attacks have been proposed; they are discussed in section IV-B. Centralized training and deployment is the current standard scheme in ML. The depicted decentralized training scheme corresponds to (horizontal) federated learning; the depicted decentralized deployment scheme is often called "Edge AI." (For decentralized, e.g., federated, training, a central coordinator is not strictly necessary: completely decentralized approaches have been proposed as well.)

### 1) CLASSICAL CYBERSECURITY

Classical cybersecurity challenges include bugs and vulnerabilities of the ML healthcare application's software stack. Such vulnerabilities can lead to a completely compromised system and massive incidents, as a growing number of security incidents show [334], [335]. As medical devices are increasingly interconnected, they become easier targets, as reports on the vulnerability of medical devices show [336].

As a result, cybersecurity now plays a major role in the approval process of medical devices [337].

### 2) CONFIDENTIALITY OF THE TRAINING DATA

Recent work [163] shows attack strategies for extracting information about training data from ML models; thus, the ML model itself breaks the confidentiality of the training data. For models that are solely used by authorized personnel (e.g., hospital personnel), such extraction attacks are not a concern. In the context of Edge AI, however, the ML model is deployed to end-users' smartphones, and any malicious party can query the model and extract information about the training data. Even ML models hosted in a protected environment and query-frequency-limited seem to be vulnerable to privacy attacks. So far, there is no indication that restricting the frequency of queries would alleviate these attacks. These privacy attacks do, however, have varied success on applications with different degrees of complexity. Recent work has shown [164] that it is harder to protect applications with high-dimensional input spaces, such as images or signals, compared to applications with low-dimensional input spaces, such as structured data, questionnaire responses, or tables with a small number of attributes. Aside from the *model* potentially leaking sensitive patient data, it is also becoming increasingly apparent that publishing medical datasets in a matter that prevents the re-identification of supposedly anonymized health data is a highly challenging endeavor. To provide two recent examples, Rocher *et al.* [100] demonstrate that using publicly available, supposedly anonymized datasets, a large fraction of the U.S. population could be re-identified (thereby defeating the purpose of anonymization). Similarly, Packhäuser et al. [119] demonstrate that given a chest X-ray recording of a patient, other recordings of the same patient can be identified among more than 100,000 recordings in a publicly available dataset with very high accuracy, thereby enabling re-identification of the supposedly anonymized recordings in the dataset. Both authors thus call for a reconsideration of the classical anonymization strategies, i.e., de-identification and population sampling.

### 3) CONFIDENTIALITY OF THE QUERY DATA
Sensitive patient data are threatened not only during model training but also when querying an ML model. In Edge AI applications, where the model remains on the patient's side while querying it, the patient's query data — serving as inputs to the model — are protected. At the same time, however, any potentially malicious party can freely query the model and thus extract potentially sensitive information about the training data (see above). If the ML model itself is an asset or simply too large for a client device (such as GPT-3 [338]), a patient would remotely access the model, and the model might even run on a cloud service. In this case, a naïve implementation could leak sensitive patient data during each query.

### 4) CONFIDENTIALITY OF THE MODEL
The model itself can be of high value. From a business perspective, it might be desirable to keep the precise model private. While classic security measures may prevent direct model access, a recent line of work [339] shows that by using a sufficiently large amount of queries, attackers can extract a model from inference only. This attack scenario becomes relevant once an attacker has either a device on which the valuable ML model is deployed or unlimited remote access to a valuable model.

### 5) INTEGRITY OF THE TRAINING PROCESS
For privacy and regulatory reasons, it may be desirable to train an ML model in a distributed manner. Federated and distributed learning both are techniques using which several parties can keep their (confidential) data local yet train a global model together [140], [340], and many groups have advocated for the use of such techniques for healthcare data [101], [141], [142], [165]. In these distributed learning settings and, more generally, if the data collection process cannot be fully trusted, the integrity of the training process has to be protected. Hitaj *et al.* [341] have demonstrated that malicious participants to the training process can extract the data of other participants, and recent work has shown [145] that malicious training parties can also manipulate the resulting global model. These manipulations can even result in backdoors: for specific inputs, the globally trained model may return maliciously crafted undesirable results, rendering the model unreliable and potentially unsafe.

### 6) INTEGRITY AND ROBUSTNESS OF THE MODEL
Beyond the integrity of the training process, ML applications face the challenge that classical ML models can — infamously — be confused by maliciously crafted inputs, called adversarial examples [168], [218], [219], [270]. Adversarial examples can be crafted from legitimate inputs with minuscule modifications that are practically invisible for a human but significantly alter an ML model's output. Naturally, in protected environments — such as a hospital — where only authorized personnel uses an ML model, adversarially crafted inputs are not a concern. For Edge AI applications and remotely accessible ML models, however, adversarial examples can render an ML model unreliable. This might concern, e.g., MML systems intended for home use. The vulnerability to adversarial examples is higher in applications with high-dimensional input spaces [169], in particular those applications where the inputs have a natural source of perturbation, such as camera images or physiological measurement signals. Humans have a high tolerance for signal variation in such settings, whereas non-robust ML models can drastically change their output if the input is only slightly modified. As an example, minimal modifications in the lighting of an image will result in a technically different image that appears indiscernible from the original image for humans but may be classified differently by a non-robust ML model. Changes in *low*-dimensional input spaces (e.g., structured data or tables with a small number of attributes) on the other hand can be spotted easier by humans and leave less freedom for unobservable modifications; an effect sometimes referred to as the curse of dimensionality in adversarial examples [169]. Finally, notice that a model's vulnerability to adversarial examples is closely related to its *robustness*, also cf. the corresponding discussion in section III.

## B. MEASURES TO ACHIEVE PRIVACY AND SECURITY

For classical *cybersecurity*, there is a rich body of literature regarding best practices, such as running known attack strategies against the software stack (so-called penetration testing), formal verification, and architecture design that isolates as many system parts as is feasible, and organizational measures, such as incidence response teams and red teaming. For an actionable summary of practical best practices for facing classical cybersecurity challenges, refer to, e.g., the NIST[16] and BSI guidelines.[17]

For ensuring that a model does not leak information about the data it was trained on, a young body of literature [166], [167] on *privacy-preserving machine learning* (PPML) presents methods to train machine learning models in a way that ensures that no information about individual patients can be extracted; an approach called *differential privacy*. As such PPML methods try to approximate the original learning task in a well-generalizable and privacy-preserving manner, these methods inherently lead to a loss in accuracy. This loss of accuracy is manageable for convex optimization problems, such as linear or logistic regression, SVMs, or rank pooling. For non-convex methods such as neural networks, however, current (differential privacy) approaches lead to a significant loss of performance [166], although this loss can be circumvented under specific circumstances [342]. In some cases, pre-training an ML model on public data and then fine-tuning this pre-trained model via privacy-preserving methods—a variant of *transfer learning*, which was already discussed in more detail in section III-B—might represent an attractive alternative from both a privacy and an accuracy perspective. To validate an employed technique, penetration testing frameworks such as the adversarial robustness toolbox (ART) [343][18] can be utilized for assessing the vulnerability of the trained model to extraction attacks, among other threats.

Distributed learning methods (see the bottom left panel in Figure 4) such as *federated learning* [139], [140] and *split learning* [302] have already been discussed in more detail in section III. They represent attractive methods for increasing the pool of available training data without having to share these data between different institutions. This is especially relevant in the face of increasingly prominent de-anonymization attacks on publicly available datasets [100], [119]. It is, however, important to note that they still suffer from the problem of information leakage from the trained *models*. To solve this problem, distributed learning methods can be combined with, e.g., differential privacy schemes, as discussed above, or homomorphic encryption [102], [103], [140], [165]. Protecting the integrity of the *training process* during a distributed learning process against malicious actors is a very young challenge that is most relevant in scenarios in which there is a large number of potentially unknown participants. The best automated techniques use different kinds of outlier detection methods that look for suspicious behavior of single training parties [344], similarly to the outlier rejection techniques discussed in section III-C. Potentially, slightly noising the model's responses might increase the number of queries that are needed to accurately extract the model [345]. These techniques increase the difficulty of injecting backdoors yet cannot provide security guarantees against strong attackers. A taxonomy and overview of different attacks on distributed learning processes and possible defenses against them can be found in Lyu *et al.* [146]. As was mentioned above, many of such integrity problems can be circumvented by limiting the number of training process participants to a relatively small, known group of institutions, such as a set of hospitals.

One way to protect the *query* data, i.e., sensitive patient data serving as inputs to the model, is to deploy the trained model in the "edge", i.e., on a patient or hospital device, thereby alleviating the need to communicate the query data with any external entity. (This corresponds to the setting in the bottom right panel of fig. 4.) However, as discussed above, if the data was sensitive, so is the trained ML model. If solely authorized personnel is involved, i.e., in a hospital setting, this is not a problem and may be no need for special protection. If, however, the ML model is accessible to unauthorized query issuers—as is common in large scale applications—, classical (not privacy-preserving) machine learning techniques can lead to undesired information leakage about the sensitive training data, as discussed above. Again, differentially private machine learning can be used to avoid such leakage. Another challenge in this setting concerns the validity of the model: if a remote server distributes the model, how can its validity be verified on an end-user device? This problem can be addressed using standard techniques for verifying remotely distributed software, as are used by, e.g., various software package managers and update mechanisms.

In the alternative scenario in which patients *remotely* access the model (which runs on a centralized server), recent work [346] has shown that trusted computing platforms (such as Intel SGX [347]) can be used to ensure that the model owner cannot access the sensitive patient query while computing the response to the query. Additionally, cryptographic solutions exist where both parties either compute on encrypted data via homomorphic encryption [348], using secure multi-party computation [349], or using outsourced computation techniques [350].

Regarding the preservation of the confidentiality of a trained model against *model extraction* attacks, the research literature is also very young. As the current attacks require a large number of queries to the model, the best countermeasure currently seems to be to limit the query rate to a model [351].

Safeguarding the integrity of an ML model against adversarial attacks, a property called *adversarial robustness*, is another young and challenging research area. This is, again, most relevant in Edge AI settings, in which model inputs can be influenced not only by certified personnel. Many so-called *best-effort* defense mechanisms, i.e., mechanisms which do not provide security guarantees, have been proposed in the literature to defend ML models against adversarial examples. Several of them, however, turned out to be unsafe against adaptive attackers who are aware of the deployed defense mechanism and its implementation [352]–[354]. See [218], [353]–[355] for an overview of defense mechanisms that are nullified by adaptive attacks. A best-effort defense mechanism is still considered practically valuable is adversarial training [168], [356], [357]. On the other hand, there are adversarial defenses that offer certifiable robustness against a subset of adversarial attacks, such as the mechanisms described by Lécuyer et al. [216] and Lécuyer *et al.* [215]. For a given input $x$, these defenses can certify for a certain distance $r$ (which depends on $x$) that no adversarial example exists within a ball of radius $r$ centered at $x$. Alas, for some inputs $r$ might be equal to zero, providing no guarantees for these inputs. An attacker restricted to such a threat model, even if adaptive, will not be able to create adversarial examples within those boundaries. Recently, Zhang *et al.* [298] have provided an interpretation of adversarial attacks as intentional exploits of spurious correlations learned by the model. The authors also provide a novel defense mechanism that leverages causal modeling for improving adversarial robustness. In general, it is crucial to carefully evaluate any implemented defense mechanism in order to avoid a false sense of security [358]. To perform such an evaluation in practice, pen-testing frameworks like the FoolBox [359], [360], ART [343][18], or Cleverhans [361] toolboxes may be employed. Finally, notice that methods to increase robustness against adversarial attacks do *not* necessarily increase robustness against natural distribution shifts [293].

---

[16]See https://www.nist.gov/itl/voting/security-recommendations
[17]See https://www.bsi.bund.de/EN/Topics/ITGrundschutz/itgrundschutz_node.html
[18]See https://art360.mybluemix.net/

### C. SUMMARY: PRIVACY AND SECURITY

Medical machine learning applications necessitate exceptionally high standards of privacy and security. Distributed learning models such as federated learning and split learning appear to be promising frameworks for solving the problem of medical data silos while maintaining patient data privacy [101]–[103], [141], [142]. For MML models solely accessible by authorized personnel, security and privacy concerns appear to be relatively limited. However, in scenarios with more widely distributed access, models should be protected from direct access due to privacy concerns. For MML models that are remotely accessible to a broader audience or for Edge AI applications, a limited query rate, robustness-enhancing training methods, and privacy-preserving learning methods should be considered. In particular, it is advisable to evaluate the security and privacy properties of such a model using popular penetration testing frameworks [343], [359]–[361]. Finally, in scenarios where the data collection or the training process is not entirely trusted (e.g., in some federated learning settings), countermeasures against backdoor injections may be necessary [146], such as deploying outlier detection methods [344].

## V. TRANSPARENCY AND EXPLAINABILITY

By *transparency*, following Dignum [58], we will refer to "the capability to describe, inspect and reproduce the mechanisms through which AI systems make decisions and learn to adapt to their environment, and the provenance and dynamics of the data that is used and created by the system. [. . . ] transparency is also about being explicit and open about choices and decisions concerning data sources and development processes and stakeholders." [58]. Thus, transparency encompasses the open communication of some or all factors that influence the decision-making process, including, among others, the training data and their characteristics, the selected model structure, the training procedure, performance on the training and test datasets, real-world performance across different patient groups, and more. Notice that transparency alone does not necessarily enable *comprehension* on the part of the information recipient: making the full structure and weights of a neural network openly available can be seen as fully transparent, but due to the model's opacity likely does not significantly increase recipient comprehension. Thus, for many purposes, transparency alone is not sufficient, and *explainability* must be a second goal. By explainability, we refer to "the ability to explain or to present *in understandable terms* to a human" (emphasis ours) [184].[19] In all except for the very simplest cases, there is a trade-off between the understandability and the completeness of an explanation [190], [362]: facilitating human comprehension necessitates meaningful abstraction and reduction of details, just like a medical doctor would not provide her whole scholarly knowledge as an explanation but rather present the few most important factors influencing her decision. Thus, useful explanations of complex models can never be complete, i.e., can never correctly reflect all influencing factors; a conundrum that is sometimes called the *approximation dilemma* [191].

In consulting current regulatory documents for medical devices, a clear demand for transparency or explainability of solutions for medical use is not to be found (although this appears likely to change in the near future, cf. section II-A). In Europe, the GDPR, granting a right to an explanation and a right to contest any automated decision, does require some (application-dependent and currently unclear) degree of explainability and transparency in automated decision-making systems.[20] Similarly, the EU commission's proposal for the regulation of AI systems [49] emphasizes the (albeit vague) need for transparency. On the other hand, transparency and explainability serve many practical purposes: they aid developers to create safe, robust, reliable, and fair systems [333], auditors to evaluate system performance [29], and doctors and patients to understand and potentially challenge algorithmic decisions [7]. From a medical device manufacturer's perspective, transparency and explainability support risk assessment, risk mitigation during development and deployment, and clinical validation of machine learning solutions, besides fulfilling a stakeholder need. Both the FDA's plan to advance the agency's oversight of AI/ML-based medical software [43] and the WHO's guidance [7] emphasize the role of transparency of AI/ML-based devices as an essential factor to achieve patient, clinician, and societal *trust* in AI/ML technologies. From an ethical perspective, transparency to the user is often understood as a prerequisite for human agency in the face of algorithmic decision-making [7], [363], [364]: if people are affected by the consequences of an ML system's decision, they must receive sufficient information to be able to exercise their rights appropriately and, if necessary, challenge the decision (as is also required by the GDPR [92]).

---

[19]Like Miller [183], [260], we do not differentiate between explainability and *interpretability* and use the two terms interchangeably.
[20]Cf. the discussion in section II-A.

### A. GOALS OF TRANSPARENCY AND EXPLAINABILITY: TRANSPARENT TO WHOM?

There are many stakeholders involved in the lifecycle of an MML system (cf. fig. 1), with different levels of expertise regarding machine learning in general and a given system in particular, and with different reasons to be interested in transparency and explainability of the system [191], [253], [259], [365]. Weller [253] lists a number of potential goals of transparency and explainability, including

- helping a *developer* and, more generally, the *manufacturer* understand the system and the way particular decisions are made, thus aiding the development of a safe, robust, and reliable system,
- helping a particular *user* (which may be a patient or a healthcare professional) and the *society at large* broadly understand the system as a whole and thus aiding the formation of trust or distrust in the fitness of the system for a particular application,
- helping a particular *user* understand the most important factors influencing a particular prediction, thus enabling him to trust or challenge it,
- helping an *auditor* or regulator assess the trustworthiness of the system as a whole or the way a particular decision was made, and
- helping any of the above stakeholders assess the development of system performance over time after market deployment.

Each of these goals requires different kinds of information, different types of explanations, and different means of communicating the explanation. In other words, the transparency and explainability–enhancing features of an MML system must be tailored towards a specific purpose [191], [365]: whom should they serve?

Crucially, transparency does not necessarily require openly communicating all aspects of the system to the public — this may be undesirable for various reasons, including privacy concerns regarding patient data and intellectual property concerns regarding the exact employed models [7], [109], [366]. For instance, it may be preferable to communicate critical implementation details such as model structure, training routine, and the full training, validation, and test datasets to auditors only [109], [366]. A developer, on the other hand, who is struggling with the severe task to create a system that is robust and reliable in the face of all the challenges discussed in sections III, IV and VI, will benefit from transparent access and well-tailored explanations of all aspects of the system [191], [333]. In all likelihood, the optimal level of detail differs between explanations for said developer, for a healthcare professional working with the system, and for a patient affected by its decision.

Finally, one particular goal of transparency may be to monitor the temporal development of a continuously learning MML system. In this case, transparency and explainability measures may serve to assess whether an iteratively improving system stays safe, robust, reliable, and fair at each time. This connection between transparency requirements and continuously learning systems is emphasized, e.g., by the FDA [31], [43].

### B. TECHNICAL MEASURES TO ACHIEVE TRANSPARENCY

Transparency should, wherever feasible, be pursued regarding all aspects of a model, its origin, and its performance [7], [113], [254]. To begin with, transparency regarding the (composition of the) training and test datasets [250] as well as the *process* by which these are gathered and curated [112] has been recognized as crucially important. This is hardly surprising, seeing that the robustness, reliability, and fairness of the resulting model are strongly dependent on (albeit not entirely determined by) the properties of these datasets. Frequently, samples are not representative of the target population and, e.g., tend to be disproportional sick or biased towards the local population [282]. Various exploratory data analysis techniques may be used to characterize a dataset in a meaningful way, including, e.g., data prototypes and criticisms [117], [367], [368], which characterize, respectively, dominant and rare (potentially under-represented) clusters of data. One may aim to be transparent regarding the full, raw training and test datasets, transparent regarding the way in which these data were collected, or transparent regarding the prevalence and characteristics of certain (age, diagnosis, sex, ethnic origin, recording type, and location) groups. What is the model's performance overall and within these different (training or test) subgroups? Are there significant discrepancies to be observed? This information may drive decisions concerning, e.g., the proper operating conditions under which the model is believed to operate reliably, the necessity to collect further data, or the necessity to modify the model class or training algorithm to improve the model's performance. Due to the constantly lurking threat of dataset shift between the training data and real-world data, post-market real-world performance monitoring is believed to play a key role in building trust in an MML system [31], [43], [49]. (Post-market surveillance is, of course, already a key requirement of presently existing medical device regulations.) To the clinical user, transparency regarding the data that were used to train the algorithm and their characteristics, the relevance of the different input data provided to the system, the logic employed by the system, the role intended to be served by its output, and data quantifying the device's performance are crucial indicators for assessing the credibility of the system in a certain setting [43]. Moreover, the conditions under which a system is safely applicable must be clearly communicated.[21] Popular transparency initiatives include the TRIPOD statement for reporting multi-variable prediction models for individual prognosis or diagnosis [254], the ''Datasheets for Datasets'' proposal [250], the Google model cards initiative [113],[22] the dataset nutrition label [252][23] and AI FactSheets [251].[24]

Achieving *contestability* requires a particular kind of transparency, namely, *traceability* [92]. If users (e.g., patients or healthcare professionals) are to be given the power to contest an automated decision post-hoc, i.e., *after* it has been made, the system must implement an infrastructure that enables a) retracing the way that particular decision was made and b) assessing whether this decision-making process adhered to the relevant regulations and requirements [92]. Tubella *et al.* [92] propose to address this technical challenge by implementing a traceability software infrastructure and enabling post-hoc verification of formalized requirements that must be fulfilled by a lawful (or otherwise desirable) decision-making process. (Also cf. the discussion of formal verification methods in section III.)

Finally, transparency about prediction or classification *uncertainty* is another essential requirement: how certain is the system regarding a particular prediction? The issue of proper uncertainty quantification has already been discussed in section III.

The following section will discuss measures to aid developers, auditors, and users in understanding the functioning of the trained model.

### C. TECHNICAL MEASURES TO ACHIEVE EXPLAINABILITY

Popular interest in explainable machine learning models has exploded in recent years, with government agencies across the globe demanding that machine learning systems be explainable, multiple major companies providing toolboxes to implement explainability, and whole MOOCs being launched to cover it [191], [238], [369], [370]. In the medical field, calls for the explainability of MML systems abound [5], [96], [158], [282], [284], [286]. To begin the discussion, a distinction has to be made between models that are *intrinsically* interpretable and black-box models that are made explainable through post-hoc explanation methods [191], [368].[25] Whereas explaining a non-interpretable, black-box model necessarily always requires omitting (supposedly less important) details and introduces further complexity into the system (the explanation method), intrinsically interpretable models have the advantage that their functioning can be explained to users without potentially misleading simplification [185].

The class of intrinsically interpretable machine learning models is larger than one might expect. Firstly, it includes all the classical, simple (rather low-dimensional) parametric models, decision trees, generalized linear models, and many more [368]. Various types of probabilistic graphical models such as Bayesian networks — including, in particular, causal models [186], [323] — are interpretable as well, at least up to a certain degree of model complexity. Bayesian rule lists were introduced to achieve interpretable and highly accurate models for stroke prediction [187]. *Causal* models (mainly applicable for use with structured or time-series data) are inherently explainable and provide explanations in a particularly natural form [158], [186], [188], besides bringing many other benefits to MML [138]. (Also cf. the discussion of causal models in section III-B.) Many more examples of interpretable model classes can be found in Marcinkevičs and Vogt [237]. Bridging the gap towards black-box neural networks, various interpretable yet highly flexible model classes have been proposed. As one example, *case-based reasoning* architectures leverage deep learning for visual perception yet are inherently interpretable [185], [189] and have also been employed successfully in medical imaging tasks [371] without losing accuracy in comparison to purely black-box models [155]. In a similar vein, Chen *et al.* [159] have recently proposed *concept whitening* as a method that can be used with any of the standard deep learning frameworks, and that enforces the learning of interpretable concepts in one layer of a deep neural network. The authors provide many examples illustrating their proposed concept, including a skin lesion diagnosis task. Notably, the authors did not find concept whitening to reduce the predictive capabilities of the network. For time series analysis, Sha and Wang [372] have discussed the use of an intrinsically interpretable recurrent neural network for mortality prediction based on diagnostic codes, also finding their proposed model to outperform baseline models. Similarly, Guo *et al.* [373] have proposed an interpretable LSTM model. Additional approaches to equip neural networks with intrinsic interpretability (as opposed to post-hoc explanations) include contextual explanation networks [374] and self-explaining neural networks [375].

A model can be a *black box* either because it is too complicated for humans to comprehend (this famously includes most deep learning models) or because it is proprietary (this may be the case in many medical applications). In both cases, an understanding of the model's functionality can only be extracted by examination of the model's response surface, i.e., by means of *post-hoc* methods. In computer vision, *saliency maps* [376]–[378], which visually represent the importance of each pixel for arriving at a particular prediction, are a popular post-hoc method to achieve explainability of black-box systems, with popular methods for their computation including layer-wise relevance propagation [192], Grad-CAM [193] and spectral relevance analysis [379]. Implementations are readily available and, in many cases, quickly yield information that is useful to the developer or the user [194], [237], [260], [333]. Their prevalent use, especially in the medical domain, has been criticized, however, because they can be misleading or unreliable [185], [195]. For less high-dimensional applications than visual computing, *feature importance* [368], *locally interpretable model-agnostic explanations (LIME)* [190], [368] and *Shapley values* [368], [380] all quantify the importance of different input components for arriving at a particular prediction, analogously to saliency maps for visual computing. Similarly to saliency maps, these post-hoc explanations can be misleading due to their simplifying nature [368], [374]. Very differently from the attention-based prediction explanation mechanisms discussed above, *example-based explanations* attempt to explain a prediction by providing examples that the model considers similar or dissimilar in some sense [368]. Wachter *et al.* [39] propose a general method for generating counterfactual explanations, i.e., examples that are close to the input data at hand (in some distance metric) but classified differently. The authors also discuss the example of a model predicting a patient's risk of diabetes. Counterfactual examples provide very naturally human-interpretable explanations [183], [381]. While some technical challenges are to be solved for a particular application problem [368] and various implementations have been proposed [237], the main challenge when implementing this method is that there are usually many potential counterfactual examples, and selecting the most explanatory is decisive for the utility of the explanation and non-trivial [196].

---

[21]If possible, any deviation from these conditions should also be detected automatically, cf section III-C.

[22]See https://modelcards.withgoogle.com

[23]See https://datanutrition.org/labels/

[24]See https://aifs360.mybluemix.net/

[25]Interpretability is *gradual*, i.e., there is no sharp boundary between interpretable and non-interpretable models. A white-box probabilistic graphical model with thousands of variables and connections between them may no longer be classified as ''interpretable''.

The post-hoc explanation methods discussed so far were all *local*: they attempt to explain how a black-box model has derived some particular prediction and what would need to change in order for it to reach a different conclusion. Another branch of explainable ML is concerned with finding *global* explanations of the *model as a whole*. Maybe the most straightforward method for doing so is to identify a *surrogate model* from an interpretable model class that approximates the original black-box model [368]. A more complex but similar approach has been pursued by Harradon *et al.* [382], who identify human-comprehensible concepts within a deep neural network, extract a probabilistic causal model and use this extracted (surrogate) model to generated explanations of the network's decisions. Of course, all methods identifying a global surrogate model have the obvious drawback that they may not faithfully represent various properties of the original model due to their approximative nature. As an alternative, one may, again, also consider different example-based model explanations. *Adversarial examples* [168] may serve to examine the brittleness of a trained model with respect to small perturbations. A model's predictions for *prototypes* and *criticisms* [367] may be examined to analyze whether, e.g., the model performs poorly for some under-represented groups. Finally, akin to classical linear model analysis [383], the influence of individual examples in the training dataset can be analyzed, e.g., using *influence functions* [384]. In this way, examples with a particularly strong influence on the resulting model parameters can be identified and checked for potential errors or confounding properties [368].

It is a popular claim that there is a trade-off between model accuracy and model interpretability [185], [191], [385]. This claim has been refuted [159], [185], and many counter-examples — i.e., interpretable models achieving state-of-the-art performance on par with black-box models — have been provided [155], [159], [185], [187], [294], [371], [374]. While simply reducing model complexity is likely to reduce model performance, there are often ways to constrain the model class to be interpretable while retaining flexibility in the dimensions that matter for achieving a high performance [159], [185], [191]. Currently, this may require hand-crafting an interpretable model class that is suitable to the target domain [155], [185], an effort that may be prohibitive in some cases. This might be changing, however. As an example, the recently proposed concept whitening method is directly applicable using standard deep learning frameworks [159]. Moreover, in a field where safety is crucial, data are scarce, and patient recordings differ in a multitude of dimensions, it may be desirable to sacrifice a bit of predictive performance to gain increased explainability of the employed model. An interesting hybrid approach might be to craft an MML system that consists of both interpretable and black-box components, thereby combining domain-specific knowledge with general black-box type adaption layers.

An expository review of explainability and interpretability can be found in Marcinkevičs and Vogt [237]. For a comprehensive review and taxonomy of explainability methods and their current challenges, refer to, e.g., Arrieta *et al.* [191] or Samek *et al.* [386]; for a review of explainability within the MML context, see Tjoa and Guan [260]. A useful, practical guidance document has been developed by the UK information commissioner's office (ICO) and the Alan Turing institute [98]. A number of both freely accessible and commercial tools implementing (mostly post-hoc) explainability methods is already available, such as the AI Explainability 360 toolkit [387],[26] InterpretML [388],[27] Google's "What-if Tool" [389],[28] similar tools in Microsoft's Azure,[29] and the captum python package.[30]

---

[26]See https://aix360.mybluemix.net/
[27]See https://interpret.ml/
[28]See https://pair-code.github.io/what-if-tool/
[29]See https://docs.microsoft.com/en-us/azure/machine-learning/how-to-machine-learning-interpretability
[30]See https://github.com/pytorch/captum

The field of explainability raises many fundamental philosophical and psychological questions, such as:

- What constitutes an explanation? What differentiates it from, e.g., a justification?
- How do we humans explain things to each other, and how do we mentally process explanations?
- Which different types of explanations are there?
- What makes an explanation ''good'' in a given context?

We did not touch upon these matters here for brevity's sake; the interested reader is referred to the DARPA XAI literature review [238] and Miller [183] for broad and excellent overviews of the interdisciplinary challenges related to crafting good explanations.

### D. NON-TECHNICAL MEASURES TO ACHIEVE TRANSPARENCY AND EXPLAINABILITY

Achieving real transparency and explainability necessitates thoughtful interaction with the target information recipient. What kind of information is she looking for? Which type of explanation does she consider useful? For these reasons, manufacturers should closely involve stakeholders in the design of a transparent and explainable MML system [260], [286] and carefully analyze the clinical decision process into which the MML system should be integrated [59]. Appropriate methods must be developed for *communicating* the information to the user in a readily accessible, comprehensible, clear, and, ideally, interactive way [183], [235], [238], [258]. This may include the creation of informative summarizing *labels* for MML systems, which may inform about the data used for training the model, the model type used, its performance on benchmark datasets and in validation studies, its intended use, and its limitations [43], [113], [254], [390]. It will likely also demand the innovation of novel explanatory interfaces [155], [235], [258], [259], [370], [371], [391]. Notice that *evaluating* the transparency and explainability of (different versions of) an MML system quantitatively represents a difficult challenge and is the subject of current research [184], [191], [194], [236]–[238]. Finally, medical users and patients, unaccustomed to working with and thinking about ML models and systems, will require proper training [43], [260], [261] to empower them to make informed decisions when interacting with an MML system.

### E. SUMMARY: TRANSPARENCY AND EXPLAINABILITY

Regardless of the exact degree to which they may or may not be legally required, there is a broad consensus that transparency and explainability are crucially important enabling properties to achieve safe, robust, reliable, and fair MML systems. This includes transparency regarding the data used for training a model and its characteristics and limitations, the model's real-world performance across different patient groups, the uncertainty of model predictions, explanations of the model as a whole, and explanations of the most important factors influencing a particular decision of the model. Data-based explanations of model behavior, such as model predictions for minority group prototypes, adversarial examples, and influential examples, may help assess the model's robustness and reliability and inform continuous model improvement. In a similar vein, the generation of counterfactual examples represents an attractive explanatory mechanism that is applicable in many domains. Post-hoc explanation methods, such as saliency maps and LIME, can be used to explain the decisions of otherwise opaque models. Due to their approximate nature, however, they can be inaccurate and misleading and hence should only be used with great care in the medical domain. It has been demonstrated in multiple studies across various domains that intrinsically interpretable model classes can achieve the same performance as black-box models; such models should thus be preferred where possible. Finally, explanations must always be tailored to a particular stakeholder group and use case. Explanation design is a highly interdisciplinary endeavor that should involve the relevant stakeholders from the beginning.

## VI. ALGORITHMIC FAIRNESS AND NONDISCRIMINATION

Due to the proliferation of machine learning systems making decisions with a significant impact on human lives, recent years have seen a surge of interest in *algorithmic fairness* research [198]. Examples of biased machine learning systems eliciting serious consequences have been widely reported in the media, including

- publicly used face recognition algorithms performing best for White men [392],
- an algorithm used to assign U.S. healthcare resources to patients discriminating against Black patients [14], and
- disparities in true positive rates of chest X-ray diagnosis systems between patients of different sex, age, race, or insurance types [13].

In a recent survey of industry ML practitioners [393], 49% ''reported that their team had previously found potential fairness issues in their products. Of the 51% who indicated that their team had not found any issues, most (80%) suspected that there might be issues they had not yet detected, with a majority (55%) reporting they believe undetected issues 'Probably' or 'Definitely' exist.'' Ethics, nondiscrimination law, and technical accuracy and reliability requirements all demand that medical machine learning (MML) systems treat patients fairly and do not create or reproduce biases [7], [40], [41], [49]. But what exactly constitutes a ''fair'' MML system, and why are ''unfair'' biases so prevalent?

### A. OBSTACLES TO ACHIEVING FAIRNESS

Fairness is a context-dependent social construct, and thus, there is no universally accepted and applicable definition of fairness in medical ethics [394]–[397]. Many different, conflicting definitions of algorithmic fairness have been proposed for general [398] as well as specifically medical [232], [397] applications of machine learning. It is readily apparent that different medical applications demand different definitions of fairness: while a healthcare resource allocation algorithm probably should not allocate resources based on sensitive attributes such as gender, ethnicity, or social status [14], a disease or treatment prediction algorithm probably should take at least some sensitive attributes into account [201], [397]. Moreover, diagnostic performance for one patient group must not be sacrificed in order to achieve fairness by means of equal(ly bad) performance across groups [130]: instead, the aim must be to achieve optimal performance for each individual group [201]. A related challenge concerns the difficulty of identifying the groups that are relevant for fairness considerations in the first place [283], [393]. These groups depend on the application at hand as well as societal norms and may not be obvious from the beginning; e.g., native speakers and non-native speakers may form fairness-relevant groups in applications related to text generation and understanding, but not in many others [393]. There is a particular risk to overlook potential sources of unfair discrimination due to blind spots in the development team [393]. For these reasons, the decision for a particular definition of fairness to pursue in a given application should generally not be made by the ML development team alone but by a broad and inclusive group of stakeholders — including, in particular, medical professionals, medical ethicists, and patients [395], [399], [400]. Given a fairness definition to aim for, there are, however, still considerable challenges involved in creating a model that conforms with this definition.

From a technical point of view, none of the negative examples mentioned above are particularly surprising: machine learning models, to a large degree, reflect the data they are trained on — and if these data are biased, then the resulting machine learning model is also likely to be biased. There are many different potential sources and types of bias in a dataset, cf. Srinivasan and Chander [401] for a recent taxonomy. As an example of *historical bias*, the health risk score algorithm studied by Obermeyer *et al.* [14] used past health care cost data as a proxy for health care needs — and since Black patients historically received inferior treatment, the algorithm assigned a lower risk score to them compared to White patients with equally severe symptoms. This example also illustrates the *measurement bias* that may arise from the incautious use of proxy variables for training ML models, such as healthcare costs, hospital visits, or medication usage as proxies for patient health. Measurement bias may also arise due to discrepancies in diagnostic and treatment recommendation accuracy across, e.g., different gender, ethnic, or weight groups [402]–[405]. Another challenge is posed by *representation bias*, i.e., groups being represented to different degrees in a dataset [120], [392]. In a study of chest x-ray disease classifiers trained on datasets with varying male–female ratios, Larrazabal *et al.* [115] found that classifiers generally performed worse for the underrepresented group. Unfortunately, many biomedical databases are known to significantly under-represent large parts of the population [111]. Finally, *biological differences* may simply render the prediction task easier or harder in different groups [406]. For example, in another study on chest x-ray disease classifiers, Larrazabal *et al.* [13] found that classifiers universally perform worse for females compared to males, despite similar numbers of male and female subjects being contained in the dataset.

The biased data problem is very hard to solve because learning algorithms often amplify even subtle correlations in superficially balanced datasets, an effect known as *bias amplification* [121], [297], [407]. Further complicating the problem, even an ideal training dataset devoid of any biases does *not* guarantee that the trained ML model will be bias-free [407]: model underspecification [299] and misspecification [408] may still introduce arbitrary biases. Hooker [407] discusses the impact of seemingly innocent technical measures and design choices on the fairness of the resulting model, including privacy-enhancing techniques, quantization, compression, as well as the choice of the learning rate and training length. (The reason that these small changes may have a large impact on the trained model lies, of course, in the model's underspecification.) To summarize, one can broadly distinguish two important technical causes for algorithms to be unfairly biased: biases present in the training data and biases introduced by the learning algorithm or the model [198], [407].

### B. DATA COLLECTION AND CURATION MEASURES TO ACHIEVE FAIRNESS

Bias is pervasive in medical datasets [276]. Electronic health records are known to be biased due to their dependence on the patients' interactions with the healthcare system. As an example, Agniel *et al.* [277] found that the timing of laboratory tests had higher accuracy in predicting three-year survival than the actual test result. Similarly, biases due to national healthcare policies [278], [279], funding systems [281], and patient distance to treatment centers [280] have been demonstrated, while studies on the effect of the implicit racial bias of healthcare workers on clinical decision-making have yielded mixed results to date [409], [409], [410], [410]. There is also a particularly subtle form of bias to be expected once machine learning systems are used for clinical decision making: the algorithms used to make a decision may be biased, which then introduces these biases into the dataset used to train the next machine learning model [231], [411].

Given that observational medical databases are almost surely biased in one way or another, how can one hope to obtain a representative, ideally unbiased dataset? There are two primary solution approaches: performing targeted data collection so that the likelihood of bias occurrence is minimized, or using synthetic data to either augment or completely replace human data. Suitably targeted data collection has been demonstrated in many applications to improve not only the fairness of the resulting classifier (e.g., by improving performance on minority groups) but also its overall accuracy [122], [412]. Aiming for a balanced dataset will typically involve targeting similar ratios of different genders, ethnicities, age groups, sensor types, and device manufacturers, but there are many more factors to consider depending on the particular application. Importantly, purely observational studies are often ill-suited for gathering a balanced dataset because of the biases introduced by the existing healthcare system, cf. above. Best practice guidelines on how best to gather balanced, representative datasets are currently sorely lacking [393], although there are example projects where this has been or is being attempted [111], [297], [412]. Unfortunately, in many medical applications, it is impractical or even impossible to gather a sufficiently large and balanced dataset. In these cases, synthetic data generation methods, such as data reweighting schemes, artificial data modifications like cropping or rotation, and pathophysiological simulation models, represent an attractive alternative. These methods have been discussed in more detail in section III-B. In a recent publication, Zietlow *et al.* [130], with the explicit goal of mitigating group performance disparities in computer vision applications, have proposed a method that adaptively performs data augmentation for disadvantaged groups, using GANs and SMOTE for synthetic image generation, yielding significantly improved worst-case performance.

Practice shows that dataset curation is very likely to be an iterative process, where some form of bias is detected either during testing or in the field, and as a remedy, the training or test datasets are adjusted to be unbiased in the observed regard [393]. However, while careful curation of the training and test datasets has been recognized as a key tool for fighting bias [393], [413], it is unlikely to fully remove all biases, and thus a successful bias mitigation strategy is likely to combine data curation techniques with algorithmic measures.

### C. ALGORITHMIC MEASURES TO ACHIEVE FAIRNESS

Proceeding to the algorithmic level, how can the likelihood of learning an unbiased model be increased? The first class of solution approaches is called *fair representation learning*, popularized by Zemel *et al.* [207]. The main idea is to find an intermediate representation of the data from which protected attributes can not be recovered but which retains the remaining information. Note that this is a much stronger requirement than merely removing protected features from a dataset because these can often still be recovered utilizing their correlations with other features [414]. The resulting sanitized dataset is then used as the input data for the actual prediction model. Recent work on fair representation learning is usually adversarial in nature, i.e., the correct representation of the original data samples is maximized while minimizing distributional differences between protected groups in the learned representation [415]. Fair representation learning is particularly amenable to the situation where a dataset is to be provided to third parties, and one wants to ensure that the learned models will likely be fair [208], [209]. If there are no third parties involved, fair representation learning may not be an appropriate solution [209].

Further advancing along the learning pipeline, the choice of an appropriate model structure clearly influences the likelihood of identifying a biased model. If a model correctly captures the potential sources of bias in a dataset, their relationships to different attributes, and the factors that can and cannot reasonably influence the prediction, it is less likely to be biased. *Causal models* allow for an accurate description of these relationships and have therefore been proposed as a promising means to learn fair models from biased datasets [204], [205], [400]. This approach is not limited to simple white-box models based on prior knowledge, as various complex and highly flexible model structures that nevertheless incorporate assumptions about causality have been proposed in the literature [204], [416]. Besides enabling the learning of fair models from biased data, causal models have various other advantages for reliability and explainability, cf. sections III-B and V-C.

Agnostic to the choice of a particular model structure, there is a large body of research on *algorithmic fairness* [198]. In this field, the general approach is to optimize a selected definition of fairness during model training. Popular fairness metrics include *statistical parity* (or *anti-classification*), where classification must be independent of protected attributes, *equalized odds* (or *classification parity*), which requires that common measures of estimator performance are equal across groups, and *calibration*, which requires that conditional on risk estimates, outcomes are group-independent [199], [233]. Many of these fairness metrics, as well as methods for enforcing them, are implemented in the open-source AI Fairness 360 (AIF) [417][31] and fairlearn [418][32] python packages; some of these metrics are also implemented in, e.g., the TensorFlow Fairness Indicators[33] and Google's "What-if Tool" [389][28]. It has been demonstrated that there are inherent problems with this approach, however. Chouldechova [233], [419] demonstrate that most of these fairness definitions are incompatible with one another, i.e., satisfying one fairness constraint implies violating another. Moreover, Pleiss *et al.* [420] demonstrate that these fairness definitions are in many ways not sufficient, and optimizing for one of these fairness metrics will often even lead to worse outcomes for members of minority classes [130], [210]. For this reason, especially concerning medical risk score models, it has been argued that the main aim should be the development of models that are highly predictive and well-calibrated across groups [210], [233]. In addition, fairness debates should consider *outcomes* instead of just predictions [210], [234]: what is the impact of deploying a given MML model on different groups? As one example, Pfohl *et al.* [210] consider the *net benefit* (i.e., the expected utility) of deploying a clinical risk score model for different patient populations. Finally, the centrality of categorizing patients into distinct groups to the fairness debate has been criticized as in reality, group assignments are often fluid, categories are based on contestable social assumptions, and patients might not (self-)identify with either of the available categories [421], [422].

---

[31]See https://aif360.mybluemix.net/
[32]See https://fairlearn.org/
[33]See https://www.tensorflow.org/tfx/guide/fairness_indicators

Recently, reflecting the general rise of the causal inference field [156], [324], various *causal* fairness definitions have been proposed [206], [235], attempting to solve the problems with the previously proposed statistical definitions mentioned above. Again, there are many different causal fairness notions, the most general of which is path-specific counterfactual fairness (PC fairness) [206]. The key advantage of these causal notions of fairness is that they can correctly adjust for confounding factors in the data due to, e.g., historical bias. Recently, Yan *et al.* [235] have developed an interactive tool that identifies the causal relationships learned by an ML model and enables users to interactively explore potential sources of unfairness in a dataset or model. The tool was favorably evaluated compared to the AIF tool, with users mentioning the increased transparency of the identified causal graph and the interactivity of the tool as key reasons [235].

One interesting challenge concerns the question of *classifier decoupling*: if a classifier trained on the whole training dataset performs better on some groups than others, (under which circumstances) would it be better to train decoupled classifiers for the different groups? This question has received significant attention in the research community [199]–[203], with various groups providing theoretical analyses and practical algorithms for deciding when it is beneficial for members of a certain group to be provided with a classifier trained just on their data versus one trained on the whole dataset. A central theme in this line of work is that of *preference guarantees* [201] (or benefit-of-splitting [202]): the majority of each group should prefer the classifier assigned to them over a) a pooled classifier that does not differentiate between groups and b) the classifiers assigned to all other groups. In this context, the challenge of subgroup fairness should also be mentioned: a classifier that appears perfectly equitable concerning two separate sensitive attributes (say, gender and ethnicity) can be arbitrarily inequitable for subgroups defined by the combination of the two attributes (say, Black women). This phenomenon is known as *fairness gerrymandering* [229] and is closely related to the problem of *inframarginality* [230]. Thus, the selection of (sub)groups with respect to which to evaluate fairness metrics is of high importance for evaluating the fairness of a model [423], [424].

A number of *post-hoc methods* have been proposed to recalibrate or fine-tune an existing (biased) model in order to satisfy some fairness metric [197], [425], [426]. These methods represent an attractive solution in domains in which retraining a whole model is very costly, and many of them are available in the AIF tool [417]. They do, however, suffer from the same problems as the other, previously mentioned, methods that optimize a global model for some fairness definition: the choice of an appropriate fairness definition is nontrivial, and optimizing for a fairness metric often leads to reduced performance on already disadvantaged groups [130], [210], [420].

Against this backdrop of negative results on fairness-constrained learning, the recognition that fairness and model *robustness* are very closely linked has recently gained traction [424], [427]–[429]. This makes intuitive sense, of course: a model that robustly generalizes to new datapoints will, loosely speaking, also perform reasonably well on various population subgroups. Thus, it maybe does not come as a surprise that various groups have argued that the best way to achieve optimal performance across subgroups is to perform standard (negative log likelihood-minimizing, unconstrained) machine learning on a representative, unbiased, and reasonably balanced dataset [130], [210], [329], [420]. Moreover, the use of different robust optimization schemes has been advocated for improving model fairness [424], [429].

### D. SUMMARY: ALGORITHMIC FAIRNESS AND NONDISCRIMINATION

Unfortunately, creating a fair ML model requires more work than picking an algorithmic fairness definition from the literature and choosing an appropriate learning procedure. Most practical problems are characterized by particular biases and complexities that require special care and are often not foreseeable in the early stages of a project [393]. There is no simple technical fix for the fairness problem [430]; the only real solution is to be aware of the problem in all stages of the development process, to be on the lookout for potential biases, and to gather feedback from diverse stakeholders. In a recent field experiment, dataset representativeness was found to be the strongest factor influencing the fairness of the resulting model [413]. On the algorithmic level, the causal lens currently appears particularly promising for learning fair models from biased data [204], [205], [295], [400] and evaluating the fairness of arbitrary models [206], [235]. Overall, a (reasonably) fair machine learning model will likely be a result of targeted data collection, resampling methods or synthetic data generation, a fairness-aware learning method, and monitoring for potential biases throughout the ML lifecycle. Moreover, achieving algorithmic fairness will often be an iterative process [235], [393]: data are collected and curated, a model is selected and trained, and the model's fairness is evaluated using one or (ideally) multiple of the metrics discussed above. The model is found to be unfair in some way, adjustments to the dataset, the model, or the training method are made, and the procedure is repeated. There is an evident close connection to transparency and explainability here: fairness metrics and interactive tools for exploring the relationships learned by a model represent transparency and explainability tools tailored towards investigating the fairness of the learned model.

Algorithmic (un)fairness is a topic that will almost certainly increase in importance as machine learning systems further permeate all areas of society. Preventing biased algorithms requires a significant investment of resources and procedural support, both of which are currently not widely experienced by industry ML practitioners [393]. Development teams' blind spots may make it harder to detect (and mitigate) fairness risks early in the development process [393]; a problem which can be partially remedied by increasing diversity and true inclusion [30], [110]. Unfair algorithms are already perceived by the public as one of the major problems with AI systems [431], and this perception is likely to be reinforced by future high-profile examples of unfair algorithms. It is thus in the interest of MML developers and manufacturers to be extremely cautious not to introduce unwanted biases in their systems. This need has recently also been noted by the FDA, which "recognizes the crucial importance for medical devices to be well suited for a racially and ethnically diverse intended patient population and the need for improved methodologies for the identification and improvement of machine learning algorithms. This includes methods for the identification and elimination of bias, and on the robustness and resilience of these algorithms to withstand changing clinical inputs and conditions." [43]. Current standardization efforts aim to support this goal by providing a comprehensive taxonomy and procedural framework for developing systems which do not exhibit unfair biases [211], [212].

## VII. CONCLUSION, IMPLICATIONS, AND FUTURE DIRECTIONS

Medical machine learning (MML) applications must adhere to the same standards as classical health or medical device software. Medical device regulations demand following best practices regarding quality and risk management, performing comprehensive verification, validation, and post-market surveillance, as well as the use of state-of-the-art risk control and mitigation techniques. In particular, potentially unreliable, unsafe, insecure, not privacy-preserving, nontransparent, unexplainable, or discriminatory MML systems pose *risks* (to the patient, the hospital, the society at large) that must be mitigated by the use of appropriate technical counter-measures. An ecosystem of technical standards on medical device software development details best practice development procedures that apply to MML systems as well, although none of these standards are currently specifically tailored towards ML applications. In addition, nondiscrimination law and data privacy regulations such as the GDPR impose further regulatory requirements [39], [41], [42], [99] To summarize, existing regulations already demand some degree of safety, robustness, reliability, security, privacy, explainability, and fairness of MML systems. In the future, foreseeable regulatory changes and standards under development will further substantiate these requirements [45], [46], [49].

Much of what is often called "responsible machine learning" [21], [61], [399] is thus — at least to some degree — already required by existing regulations. A notable exception from this pattern concerns *algorithmic impact assessments*; especially ones with a scope surpassing the impact on the immediate *data subject* (often: the patient) and including, among others, the impact on the hospital and its staff, the healthcare system, and society as a whole. Such broad algorithmic impact assessments appear highly desirable from a responsibility perspective [7], [33], [59], [64]–[66], [97], [240] but are currently not required by regulations.

Unfortunately, there is currently a lack of precise regulatory guidance and technical best practice documents.[34] This lack is not coincidental, however. Practically all of the technical challenges discussed in this document are extremely young, and the theory and research fields devoted to their solution are still emerging. Compared to classical, handwritten health software, creating an ML model that satisfies various desirable requirements (and verifying that it really does) can be much more challenging due to the inscrutability of the training process, its complex dependency on the training data, and the complexity and opaqueness of the resulting model. Moreover, model underspecification [299], data scarcity [1], [282], [284]–[286], distribution shift [290]–[293], and spurious correlations [10]–[12], [116], [297] pose serious challenges to overcome. They are very difficult to solve with black-box models that do not employ any domain knowledge and i.i.d. testing, as evidenced by the increasingly large number of problematic incidents with ML systems [9]. While there is at the moment considerable ambiguity regarding the exact regulatory requirements as well as the auditing of said requirements by notified bodies [69], it is evident that MML developers and manufacturers must not neglect these crucial challenges. Several key strategies towards their solution emerge. To summarize the following discussion, fig. 5 illustrates several key problems that may arise using a classical ML workflow, as well as potential solutions to these challenges, and table 1 provides a complementary summary in tabular form.

First, the careful curation of large, realistic, and representative *datasets* is indispensable to achieve robust, reliable, and fair MML. In many instances, data curation will be an iterative process: once a problem with a model is noticed, appropriate example data are added to the training data or faulty and misleading examples are removed, and the model is retrained using this extended dataset. While various initiatives are attempting to gather large, openly available, and representative datasets, data scarcity remains a key challenge for MML due to technical, organizational, and legal reasons. Policy initiatives such as the proposed European health data space are an essential step to boost the availability of high-quality medical datasets. On the other hand, distributed learning schemes such as federated learning and split learning appear very promising for enabling the training of models on large and diverse datasets without requiring these data to ever leave their respective source institutions. Due to the increased number of participants in the training process, these methods require careful consideration of privacy and security risks, however. Furthermore, transfer learning, data augmentation, automated labeling, and synthetic data generation represent additional ways to boost dataset size despite a lack of real-world (labeled) data. However, the extent to which these techniques can help (and under which circumstances) is currently unclear.

---

[34]The recent guidance by the German Fraunhofer IAIS institute [29] is, to the authors' knowledge, the first comprehensive practical guidance for developers and auditors concerning trustworthy AI.

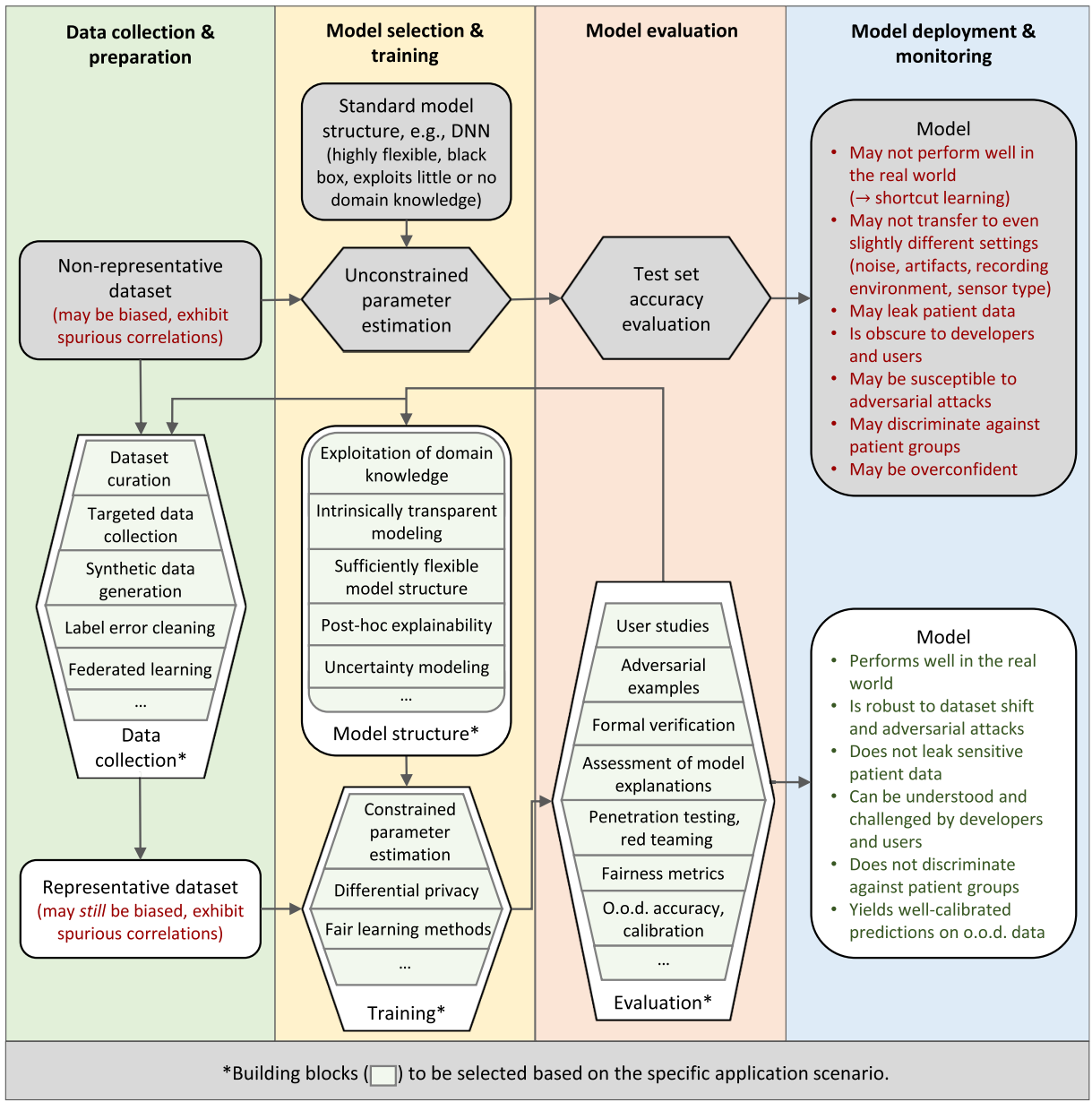

**FIGURE 5.** A visual summary of known important problems with the classical ML workflow as well as potential solutions to these challenges. The lower path is not meant to be pursued in its entirety; instead, the requirements in each specific application scenario must be analyzed carefully and the appropriate building blocks selected. Neither the list of problems nor the list of solution building blocks is considered comprehensive.

Second, the careful exploitation of *domain knowledge* increases robustness and reliability without decreasing real-world accuracy. Domain knowledge can be exploited in various ways, for example by selecting an appropriate model structure or imposing hard constraints or soft priors on the model parameters. Modeling assumptions such as monotonicity, convexity, symmetry, and smoothness may already yield significantly improved robustness and reliability while retaining flexibility in all other regards [152], [153]. The introduction of *causal* assumptions into the modeling and estimation process appears particularly promising for a number of reasons. It is purported to enable learning dataset shift-stable models [156], [295], [324], identifying fair models from biased data [204], and increasing general robustness (also to adversarial examples) [156], [298], [324] and explainability to human users [183], [381]. The causal ML field is still in its infancy, and many technical challenges remain to be solved [156], [324], but this appears to be a very promising direction for future research.[35]

Third, for assessing the generalization capabilities of MML systems, evaluating their performance on out of distribution

---

[35]The rise of causal inference has already had a significant impact on, for example, epidemiological research in recent years [432].

(o.o.d.) data, such as from a country, hospital, or time that has not been part of the training data, is essential [213], [254], [299]. This concerns both the model's predictive accuracy as well as its uncertainty estimates [174], [177]. In addition, proper evaluation methodology must be adhered to, in order to avoid exaggerated performance claims [8], [213], [331]. Inclusive consensus processes for the standardized validation, benchmarking, and reporting of MML systems may play a vital role in these regards [254], [433], [434]. Domain knowledge can be exploited for automated o.o.d. verification, by formally verifying whether a trained model satisfies application-specific constraints [220], [222] or automatically generating safety-critical corner cases for testing the model [223].

Fourth, *transparency and explainability* are not only (to some application-specific degree) legally required, but also instrumental properties that help maintain human agency in the face of algorithmic decision-making [7] and achieve robustness, reliability, and fairness of the MML system. Tools to achieve transparency and explainability include transparent communication of the dataset, its properties, influential instances and outliers, the choice of an interpretable model class, post-hoc explainability methods for generating explanations of black-box models, and algorithmic fairness metrics. If used well, they enable a developer to better judge and debug the system, an auditor to better assess the trustworthiness of a system, and a user to better understand the reasoning behind a particular decision of the system. To achieve these goals, however, transparency and explainability tools must always be tailored towards a particular user group and use case [191], [365], requiring early involvement of that user group in the development process: which information do they need? A particular kind of transparency — namely, *traceability* of individual decisions — may be required to enable *contestability* of the system, as is required by, e.g., the GDPR. Finally, care must be taken not to cause more harm than good with an explanation, since especially approximative post-hoc explanation methods such as LIME and saliency maps can be misleading [368], [374]. Where feasible, intrinsically interpretable models thus appear preferable, and such models have been proposed for many domains and have been shown to perform equally well as black-box models [155], [159], [185], [187], [294], [371], [374].

Fifth, and maybe most importantly, developing a responsible and regulatory conform MML system is not only a technical challenge; it is also an organizational and procedural challenge [7], [435]. Truly responsible AI will never be the result of simply following some "ethics checklist" [58]. It will always require virtuous designers, engineers, and managers who are truly motivated to create MML systems that benefit patients, healthcare professionals, and the society at large. Technical guidelines and requirements are no replacement for an intense occupation of all involved with the underlying ethical and technical challenges [7], [58], [59], [436]. The focus must be both on a responsible *process* as well as on a responsible end result [7], [58]. To this end, the inclusion of all relevant stakeholders is indispensable [7], [399]: ML experts, clinicians, regulators, clinical managers, patients, and relatives should be involved in the development process. Moreover, some parts of this responsible innovation process can be formalized, by, for example, invoking a red team responsible for breaking the MML system [228], standardizing algorithm audits [88], or aiming for an inclusive and transparent development process [7].

Many of the aforementioned challenges and solution approaches become significantly more complex in the context of *dynamic* MML systems. By dynamic MML systems, we refer to systems that interact with a dynamic environment in a closed loop — such as a system controlling a patient's mechanical ventilation — and systems that learn continuously over time. How can we ensure, e.g., the maintained robustness, reliability, and fairness of a closed-loop or continuously learning system? Moreover, such systems raise additional critical challenges regarding the long-term impact of deploying them. These problems are closely related to the field of classical closed-loop control theory, which has investigated the properties of dynamic systems for decades.

Lastly, the aim of this document is, of course, *not* to discourage developers and manufacturers from developing ML-based medical solutions, but to assist in doing so *responsibly* and while conforming with current and future regulations.

## ACKNOWLEDGMENT

The authors would like to thank Jannik Prüßmann, B.Sc., for assistance with the formatting of the references. Neither the funding agency nor the involved private companies have had any influence on the planning and design of this survey, the writing of the manuscript, or the publication process.

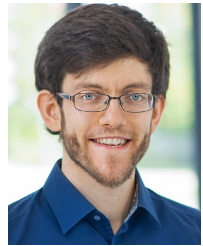

**EIKE PETERSEN** (Member, IEEE) received the B.Sc. degree in computer science and engineering from the Hamburg University of Technology, Germany, in 2012, and the M.Sc. degree in industrial education from the University of Hamburg, Germany, in 2015, having spent one semester at the Institut National des Sciences Appliquées, Toulouse, France. During the entire term of his studies, he was with Dräger Medical GmbH, Lübeck, Germany, working on signal processing problems related to mechanical ventilation. From 2015 to 2021, he worked as a Research Associate with the Institute for Electrical Engineering in Medicine, University of Lübeck. Currently, he is employed at DTU Compute, Technical University of Denmark, Denmark, investigating fairness in the context of machine learning in medicine. His research interests include the intersection of mathematical modeling, statistical inference, and the responsible application of such techniques in medicine and society.

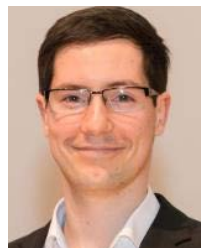

**YANNIK POTDEVIN** received the bachelor's degree in business computer science and the master's degree in computer science from Kiel University, in 2014 and 2017, respectively, where he is currently pursuing the Ph.D. degree. He is a Research Assistant with the Reliable Systems Group, Department of Computer Science, Kiel University. His research interests include robust machine learning, reliable systems engineering, and autonomous systems.

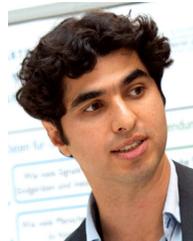

**ESFANDIAR MOHAMMADI** received the Ph.D. degree from the Department of Computer Science and Mathematics, Saarland University, in 2015. From 2016 to 2019, he worked as a ZISC Fellow at the Institute of Information Security, ETH Zürich. Since 2019, he has been a Tenured Professor with the Institute for IT-Security, University of Lübeck. His research interests include privacy-preserving approximations, in particular for machine learning, anonymous communication, and cryptographic justification of symbolic abstractions.

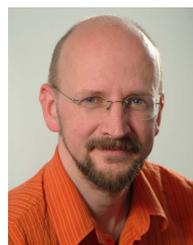

**STEPHAN ZIDOWITZ** received the Ph.D. degree in physics from the Technical University Carolo-Wilhelmina at Braunschweig, Germany, in 1997. After working in the IT industry for several years, he joined the Center for Medical Image Analysis and Visualization (MeVis), Bremen (now Fraunhofer MEVIS—Institute for Digital Medicine), in 2002. Currently, he is the Head of quality management with Fraunhofer MEVIS and is working as a Senior Research Scientist in medical image analysis and computer-assisted surgery. Since 2018, he has been working as a Lecturer on quality management and medical device safety with the University of Applied Sciences Bremerhaven, Germany. His research interests include medical image computing, statistical data analysis, and clinical decision support software. Besides these research interests, he takes an active part in transferring research results into medical products. In 2018, he was awarded the Joseph-von-Fraunhofer Prize for his work on "Algorithm for liver surgery."

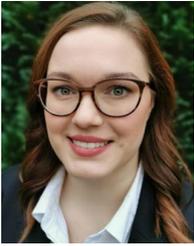

**SABRINA BREYER** received the B.A. degree in combined studies of economics and ethics/social sciences from the University of Vechta, in 2017, and the M.A. degree in practical philosophy of the economy and the environment from Kiel University, in 2020. Currently, she is a doctoral student. She is a practical philosopher and ethicist in the interdisciplinary fields of responsible research and innovation (RRI) and corporate social responsibility (CSR). Her bachelor's thesis comprises a concept for respecting human rights in business processes and throughout the supply chain. In her master's thesis, she analyzed the UN's sustainable development goals in an industrial context. Her research interests include the consideration of ethical and societal aspects of AI in innovation processes.

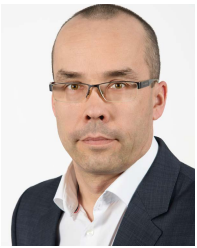

**DIRK NOWOTKA** received the Ph.D. degree from the University of Turku, Finland. He was a Postdoctoral Researcher with ETH Zürich, Switzerland; and gained his habilitation at Stuttgart University, Germany. In 2011, he joined Kiel University, Germany, as a Heisenberg-Professor, where he currently leads the Dependable Systems Group as a Full Professor. He has published more than 90 peer-reviewed articles, including ones in such high-ranking journals as the *Journal of the ACM* and the *SIAM Journal on Computing*.

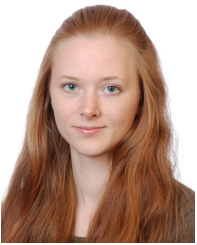

**SANDRA HENN** received the B.Sc. and M.Sc. degrees in medical engineering science from Universität zu Lübeck, in 2018 and 2020, respectively. She currently works as a Research Associate with the Institute for Electrical Engineering in Medicine, Universität zu Lübeck. Her research interests include cross-innovation topics and the safety of automated functions.

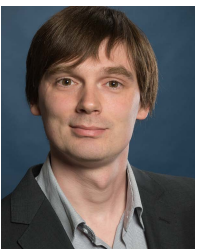

**LUDWIG PECHMANN** received the Master of Science degree in information technology from the Kiel University of Applied Sciences, Kiel, in 2013. He can fall back on many years of industrial experience. In 2019, he successfully completed a part-time training as a Regulatory Affairs Manager. Since 2016, he has been responsible for the quality assurance of the Software Department for Haag Streit Möller Wedel, Lübeck, a manufacturer of surgical microscopes; and was able to significantly improve the software quality assurance process and the product. In 2016, he successfully completed an apprenticeship as a quality management officer. Before that, he was able to work as a Software Developer and gained experience in the area of a Class III medical product at Philips Medical in software tooling. Since 2020, he has been employed at UniTransferKlinik Lübeck GmbH as a Research Assistant and played a key role in the implementation of the joint project KI-SIGS. His research interest includes the investigation and elaboration of the regulatory basis for the approval of AI-based medical devices.

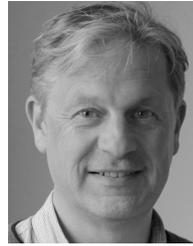

**MARTIN LEUCKER** received the Ph.D. degree from RWTH Aachen University, Germany. He worked as a Postdoctoral Researcher with the University of Pennsylvania, USA; and Uppsala University, Sweden. He pursued his habilitation at TU München, Germany. He is currently a Professor with the University of Lübeck, Germany, heading the Institute of Software Engineering and Programming Languages. Moreover, he is the CEO of UniTransferKlinik Lübeck GmbH, which also focuses on regulatory affairs, especially for medical devices. He is the author of more than 130 peer-reviewed journal and conference articles ranging over software engineering, formal methods, and theoretical computer science. His research interests include software engineering; correct and reliable systems; and formal methods and theoretical computer science, both from a technical and a regulatory perspective.

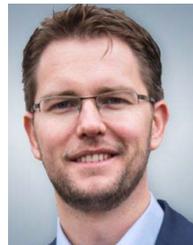

**PHILIPP ROSTALSKI** (Member, IEEE) received the Ph.D. degree from ETH Zürich, Switzerland, and served as a Feodor Lynen Scholar with the Department of Mathematics, University of California at Berkeley, USA. He is currently a Professor of electrical engineering in medicine and the Head of the corresponding institute with the University of Lübeck. Since 2020, he has been one of the directors of the Fraunhofer Research Institution for Individualized and Cell-based Medical Engineering, Lübeck. His research interests include model- and data-driven methods in signal processing and control for safety-critical systems. The primary application domains are biomedical and autonomous systems.

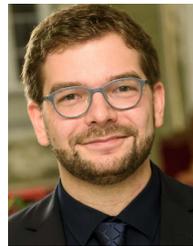

**CHRISTIAN HERZOG** (Member, IEEE) received the B.Sc. and M.Sc. degrees in mechatronics from the Hamburg University of Technology, Germany, in 2008 and 2011, respectively, the M.A. degree in applied and professional ethics from the University of Leeds, and the Ph.D. degree from the Institute of Control Systems, Hamburg University of Technology, with a focus on the robust and parameter-varying control of nonlinear and interconnected systems. He is a Transdisciplinary Researcher. Since 2015, he holds a tenured position with the Institute for Electrical Engineering in Medicine, University of Lübeck. While having engaged in the ethics and sustainability issues of technology by conducting peer-to-peer teaching since 2011, since 2016, his research interests include the ethics of engineering, innovative technologies, and artificial intelligence in particular. He commits to teaching engineering ethics, promoting broad, and inclusive participation in technology development, as well as researching and fostering interdisciplinary approaches in engineering research for the benefit of our world's ecology and society.

• • •